\documentclass{article}

\usepackage{amsmath}
\usepackage{amssymb}
\usepackage{amsthm}
\usepackage{booktabs}
\usepackage{graphicx}
\usepackage{microtype}
\usepackage{subcaption}
\usepackage{longtable}
\usepackage{multirow}
\usepackage{diagbox}

\usepackage[accepted]{icml2026}

\usepackage{enumitem}
\usepackage{xcolor}
\usepackage{xspace}

\definecolor{citecolor}{rgb}{0.0, 0.2, 0.45}
\definecolor{linkcolor}{rgb}{0.55, 0.0, 0.35}
\definecolor{urlcolor}{rgb}{0.0, 0.2, 0.55}

\definecolor{fontnonfoundation}{HTML}{298127}
\definecolor{fontfoundation}{HTML}{CC6713}
\definecolor{fontmethod}{HTML}{216091}

\usepackage[pagebackref=true,breaklinks=true,colorlinks,citecolor=citecolor,linkcolor=linkcolor,urlcolor=urlcolor,bookmarks=false]{hyperref}

\usepackage{lib}

\theoremstyle{plain}

\theoremstyle{definition}

\theoremstyle{remark}

\icmltitlerunning{\method}

\begin{document}

\twocolumn[
  \icmltitle{\method: Efficient Hyperparameter Ensembles for Tabular Deep Learning}

  \icmlsetsymbol{equal}{*}

  \begin{icmlauthorlist}
    \icmlauthor{Yury Gorishniy}{1}
    \icmlauthor{Akim Kotelnikov}{2,1}
    \icmlauthor{Ivan Rubachev}{1,2}
    \icmlauthor{Artem Babenko}{1,2}
  \end{icmlauthorlist}

  \icmlaffiliation{1}{Yandex}
  \icmlaffiliation{2}{HSE University}

  \icmlcorrespondingauthor{Yury Gorishniy}{yurygorishniy@gmail.com}

  \icmlkeywords{Machine Learning, ICML, Deep Learning, Tabular Data}

  \vskip 0.3in
]

\printAffiliationsAndNotice{}

\begin{abstract}
    In deep learning for tabular data, efficient ensembles of multilayer perceptrons (MLPs) have recently emerged as effective and practical architectures.
    Existing methods of this kind use the same hyperparameters for all underlying MLPs, which requires hyperparameter tuning for achieving the best performance.
    In this work, we introduce \method, an efficient MLP ensemble with strong out-of-the-box performance and reduced reliance on traditional tuning.
    In a single run, \method samples and trains many MLPs with different hyperparameters efficiently in parallel and selects ensemble members on the fly during training.
    Thus, \method only requires specifying ranges from which to sample MLP hyperparameter rather than exact hyperparameter values, which naturally demands less precision for good performance.
    In experiments on medium-to-large public datasets, \method\ with default settings performs on par with extensively tuned prior methods, thus substantially reducing effort and compute resources needed to achieve competitive results on tabular tasks.
    Notably, running the default \method configuration on a modern MacBook took less time than tuning some baselines on an industry-grade GPU.
    The source code is available at \href{https://github.com/yandex-research/tabpack}{this URL}.
\end{abstract}

\begin{figure}[h!]
    \centering
    \includegraphics[width=0.99\linewidth]{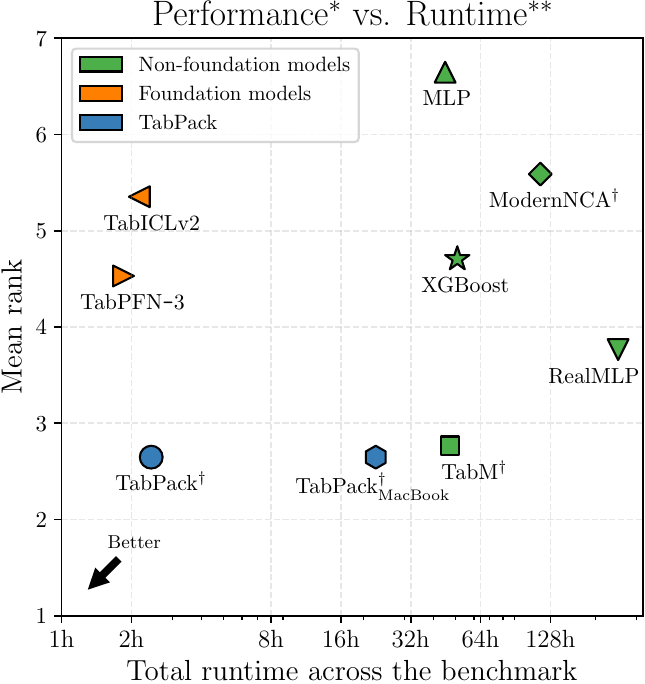}
    \caption{
        A summary of the results from \autoref{fig:performance-runtime}, obtained on \ndatasets tabular datasets spanning classification and regression tasks with up to $700$K+ training samples and up to $900$+ features.
        $\method^\dagger_\text{MacBook}$ is evaluated on Apple M4 Pro chip with 20 GPU cores.
        Other methods are evaluated on NVIDIA A100 GPU.
        \\$(^*)$ Task performance varies across benchmarks.
        For a more complete picture, see \autoref{fig:performance-runtime} (our benchmark), \autoref{A:sec:evaluation-on-large-datasets} (large-scale industrial datasets with temporal splits) and \autoref{a:sec:evaluation-on-tabarena} (small-to-medium datasets with IID splits).
        \\$(^{**})$ For
        non-foundation
        models, runtime is the hyperparameter tuning time.
        For
        foundation
        models, runtime is the inference time with default hyperparameters.
        For
        \method,
        runtime is the training time of a single run with default hyperparameters.
    }
    \label{fig:teaser}
\end{figure}

\section{Introduction}
\label{sec:introduction}

Supervised learning on tabular data is a common machine learning (ML) task in real-world applications.
For a long time, the dominant approach to such tasks was gradient-boosted decision trees (GBDTs) \citep{chen2016xgboost,prokhorenkova2018catboost,ke2017lightgbm}.
Due to the rapid progress over recent years, modern tabular deep learning (DL) models now also demonstrate strong performance and continue to improve \citep{holzmüller2024better,gorishniy2025tabm,ye2024modern,ericson2025tabarena,qu2026tabiclv2,grinsztajn2026tabpfn3}.

A common view in ML is that ensembles (mixtures of multiple models) outperform single models \citep{fort2020deep}, and tabular ML is no exception \citep{ericson2025tabarena}.
In tabular DL architectures, this idea has recently been realized in an efficient and practical form with efficient ensembles of multilayer perceptrons (MLPs) \citep{gorishniy2025tabm}.
In such architectures, many MLPs are packed together in one module by using techniques such as BatchEnsemble \citep{ioffe2015batch} or Packed-Ensembles \citep{laurent2022packed}.
This provides the performance gains from multiple predictions while maintaining reasonable training and inference efficiency.
Importantly, representative tabular MLPs are often relatively small, making it feasible to include tens or hundreds of them in a single ensemble even on one GPU.

In existing implementations of efficient tabular MLP ensembles, such as TabM \citep{gorishniy2025tabm}, the underlying MLPs have the same hyperparameters, and their diversity comes from the differences in initializations, dropout masks, training batch sequences.
To specify the MLP hyperparameters, one can either use the default values suggested by the method authors or perform hyperparameter tuning.
However, the default hyperparameters may be suboptimal for a given task, while tuning requires time and compute resources.
The lack of a universally strong starting configuration and the need for hyperparameter tuning are common among tabular ML methods, are not unique to models such as TabM, and generally make top-tier results less accessible.

In this work, we introduce \method --- a new kind of efficient MLP ensemble with strong out-of-the-box performance and reduced reliance on hyperparameter tuning, for ML tasks on tabular data.
In a nutshell, \method samples many diverse MLPs, trains them efficiently in parallel, and selects a subset with the best collective performance during training.
In particular, for the most impactful model and optimizer hyperparameters, one only needs to provide sampling ranges rather than specific values, requiring less precision to achieve good performance.
To make this approach efficient and practical, \method relies on our technique, \textit{packed hyperparameter ensembling}, which enables storing and training all MLPs with different hyperparameters within a single module.
Overall, \method strikes an appealing balance between task performance, efficiency, and sensitivity to hyperparameters that is not available with traditional approaches.

\textbf{Main contributions.}

\begin{enumerate}[nosep,leftmargin=1em]
    \item
    We present \method\ --- an efficient ensemble of MLPs with different hyperparameters for supervised learning on tabular data.
    \method packs many (e.g., tens or hundreds of) randomly sampled model--optimizer configurations into a single system, trains all models in parallel and selects ensemble members on the fly.

    \item
    On medium-to-large public datasets with up to 13.6M training instances, we show that \method\ achieves competitive performance without hyperparameter tuning.
    Remarkably, this allowed us to run the main \method experiments on a modern MacBook on equal terms with prior methods run on a discrete GPU.

    \item
    We show that, for \method, the main role of hyperparameter diversity is to enable the tuning-free workflow rather than improving performance by diversifying base models.
\end{enumerate}

\section{Related Work}
\label{sec:related}

\textbf{Tabular ML.}
Supervised ML on tabular data is typically addressed with traditional ML models or neural networks.
The strongest traditional methods are gradient-boosted decision trees (GBDTs) \citep{chen2016xgboost,prokhorenkova2018catboost,ke2017lightgbm}.
Recent DL developments include new architectures \citep{gorishniy2022embeddings,gorishniy2025tabm,ye2024modern,marton2024grande}, training techniques \citep{rubachev2022revisiting,holzmüller2024better,jeffares2023tangos}, foundation models \citep{qu2026tabiclv2,grinsztajn2026tabpfn3}.
Notably, multilayer perceptrons (MLPs) and their derivatives remain the default choice in modern non-foundation tabular architectures \citep{holzmüller2024better,gorishniy2025tabm}.
Our model, \method, is an efficient ensemble of MLPs with different hyperparameters representing a novel architectural pattern for tabular DL.

\textbf{Ensembles.}
In machine learning, ensembling refers to aggregating the predictions of multiple models.
In deep learning, a ``deep ensemble'' is a group of DL models with the same architecture trained independently \citep{jeffares2023joint} from different initializations.
Deep ensembles are known to outperform single models \citep{fort2020deep}, and the diversity of ensemble members is often viewed as an important factor behind their collective performance \citep{allenzhu2023towards}.

\textbf{Efficient deep ensembles.}
To reduce the cost of training multiple DL models, one can use efficient ensembling techniques, such as packing multiple independent models into one \citep{laurent2022packed}, sharing weights between ensemble members \citep{wen2020batchensemble,turkoglu2022film}, and others \citep{lakshminarayanan2017simple}.
Our model, \method, packs many models into one in the spirit of Packed Ensembles \citep{laurent2022packed}, with the key difference that, in \method, the individual models differ in their hyperparameters, as do their optimizers.
Conceptually, \method\ is also compatible with weight sharing among groups of members, although we do not use this technique in this work for simplicity.

\textbf{Efficient hyperparameter ensembles.}
To further diversify ensemble members, one can train models from the same family with different hyperparameters.
As with deep ensembles in general, some studies aim to do this efficiently.
One example is hyper-deep ensembles \citep{wenzel2020hyperparameter}, in which ensemble members are diversified by dropout rates and $L_2$ regularization coefficients.
Another related example is RobOD \citep{ding2022hyperparameter}, in which individual models are diversified by depth and width.
Our method, \method\, further advances this idea and shows that, in principle, one can efficiently diversify any degree of freedom in models or any part of the training pipeline, such as optimizer hyperparameters, as long as the corresponding computations can be vectorized on modern hardware.

\begin{figure*}[h!]
    \centering
    \includegraphics[width=0.99\linewidth]{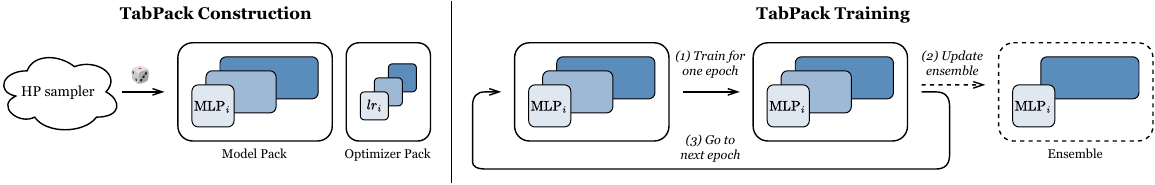}
    \caption{
        A high-level overview of \method.
        \textit{(Left)} \method is constructed by packing multiple base models and optimizers with randomly sampled hyperparameters into a single model--optimizer pair.
        \textit{(Right)}
        Training \method consists of training its base models independently in parallel and updating the ensemble once per epoch using intermediate base-model checkpoints.
        The ensemble is the main output of \method training.
    }
    \label{fig:method-overview}
\end{figure*}

\textbf{Ensembles in tabular ML.}
In tabular ML, the best performance is often achieved with ensembles.
For example, recent work on the TabArena benchmark \citep{ericson2025tabarena} highlights hyperparameter ensembles (ensembles of models from the same family) and heterogeneous ensembles (ensembles of models from different families implemented with AutoGluon \citep{erickson2020autogluon}) as strong performers.
These are examples of what we call \textit{post hoc} or \textit{offline} ensembles, where individual models are trained independently in an ensemble-unaware manner and then combined by an ensemble algorithm.
Such a post hoc approach to ensembling may not fully realize its potential in terms of task performance, and can make ensembles less efficient and less practical, particularly in production settings.
By contrast, our method, \method, represents an \textit{ensemble-first} approach, where the goal of building an ensemble is taken into account both at the implementation level for better efficiency and at the algorithm level for better task performance.

Among tabular DL ensemble models, some methods implement differentiable tree ensembles, such as NODE \citep{popov2020neural} and GRANDE \citep{marton2024grande}.
However, NODE did not outperform single models in more recent studies \citep{gorishniy2021revisiting}, and the GRANDE study is limited to classification problems, whereas our model, \method, is not restricted to any particular task type.

Another tabular DL model closely related to this work is TabM \citep{gorishniy2025tabm}.
TabM is designed to represent MLP ensembles efficiently, and it performs early stopping based on the online performance of the ensemble, not of individual MLPs.
Both points also apply to \method, however, there are several differences.
\textit{First} and most importantly, in TabM, the base MLPs share the same hyperparameters, which necessitates traditional hyperparameter tuning.
By contrast, in \method, base-model hyperparameters are sampled randomly, allowing \method\ to build powerful ensembles with little to no tuning.
\textit{Second,} \method\ selectively includes only some base models in the ensemble, whereas TabM includes all of them.
\textit{Third,} by default, TabM uses weight sharing between base models, while \method does not.
In principle, \method\ is compatible with weight sharing, and even without it, \method\ fits on modern GPUs.

\section{\method}
\label{sec:method}

In this section, we present \method\ --- a method for efficiently building powerful MLP ensembles for tabular data.
Depending on the context, ``\method'' can denote either the model itself or the overall system including the model, optimizer, and other components.

\textbf{Overview.}
\autoref{fig:method-overview} provides a high-level illustration of \method.
The key technical elements of \method\ include:

\begin{enumerate}[nosep,leftmargin=1em]
    \item
    A model architecture design in the style of Packed Ensembles \citep{laurent2022packed} for \textit{faster training} of base models in parallel (\autoref{sec:method-packed-ensembles}, \autoref{sec:method-model-pack}, \autoref{sec:method-optimizer-pack}).

    \item
    Simultaneous training of all base models for \textit{better task performance} through online ensemble construction (\autoref{sec:method-ensemble}) and early stopping based on ensemble performance as in TabM \citep{gorishniy2025tabm} (\autoref{sec:method-training}).

    \item Randomly sampled hyperparameters for the base models and optimizers making \method\ usable with \textit{little to no tuning} (\autoref{sec:method-hyperparameters}).
\end{enumerate}

The rest of this section describes \method\ in detail.

\subsection{Preliminaries}
\label{sec:method-preliminaries}

\textbf{Notation.}
This work considers supervised regression and classification problems on tabular data.
For a given object in a dataset, we use $x$ and $y$ to denote its features and label, respectively, and $\hat{y}$ to denote the prediction of $y$ by an ML model.
We use $d$ to denote dimensionalities and $W_i$ to denote linear-layer weights in the $i$-th layer of a neural network.
All notation may be used with additional subscript or superscript labels depending on the context.

\textbf{Benchmarks.}
Our benchmark is derived from \citet{gorishniy2025tabm} and includes eight industrial datasets with temporal splits from the challenging TabReD benchmark \citep{rubachev2025tabred} and nine datasets from other sources.
This gives us \ndatasets\ medium-to-large regression and classification datasets of diverse domains and sizes ranging from 10K to 1M+ objects and from 8 to 900+ features.
See \autoref{A:sec:datasets} for details.

\textbf{Experiment setup.}
We follow the experiment setup of \citet{gorishniy2025tabm} and describe it in detail in \autoref{A:sec:impl-experiment-setup}.
In particular, we split each dataset into training, validation, and test sets, where the validation part is used for things like hyperparameter tuning and early stopping, and the test part is used to compute the final metrics.
The metric definitions, namely ranks and relative improvements over MLP, are also inherited from \citet{gorishniy2025tabm}.

\subsection{Packed Ensembles: a Quick Recap}
\label{sec:method-packed-ensembles}

\begin{figure}[h!]
    \centering
    \includegraphics[width=0.95\linewidth]{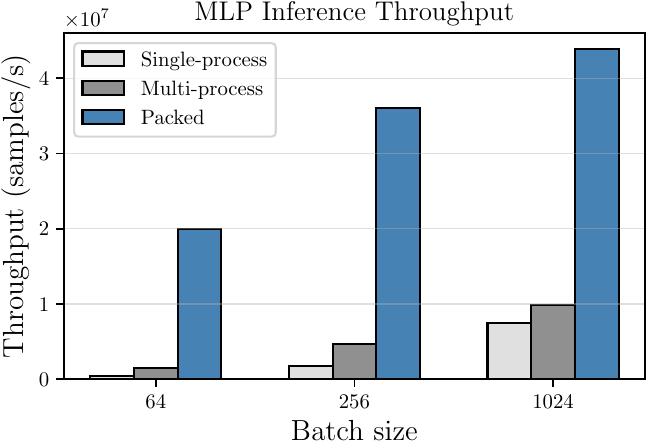}
    \caption{
        Inference throughput for an MLP of depth \defaultdepth\ and width \defaultwidth\ on one NVIDIA A100 GPU with different batch sizes in three regimes: \textit{(1)} single-process, \textit{(2)} multi-process, where multiple processes run in parallel using the same GPU (the process count is tuned to show the full potential of this approach), and \textit{(3)} packed (\nbasemodels\ MLPs are stacked into one and applied to \nbasemodels\ stacked batches in parallel).
    }
    \label{fig:matmul-throughput}
\end{figure}

Efficiently constructing ensembles from base MLP models is the central topic of our work.
To train base models as fast as possible, we use a design in the spirit of Packed Ensembles \citep{laurent2022packed}.
For our purposes, it is enough to see this method as a way to pack multiple models of the same architecture into one by simply (1) stacking the model parameters and inputs across a new pack dimension, and (2) applying all models to all inputs in parallel by relying on the broadcasting mechanism available in mainstream DL frameworks.
To give an idea of the speedups from this design, we compare the inference throughput of tabular MLPs in \autoref{fig:matmul-throughput}.
The figure covers three inference regimes corresponding to three potential approaches to training many individual models on one GPU: sequential, parallel and packed training.
The results indicate that packing is dramatically more efficient than the alternatives.

\subsection{Model Pack}
\label{sec:method-model-pack}

\begin{figure*}[h!]
    \centering
    \includegraphics[width=0.99\linewidth]{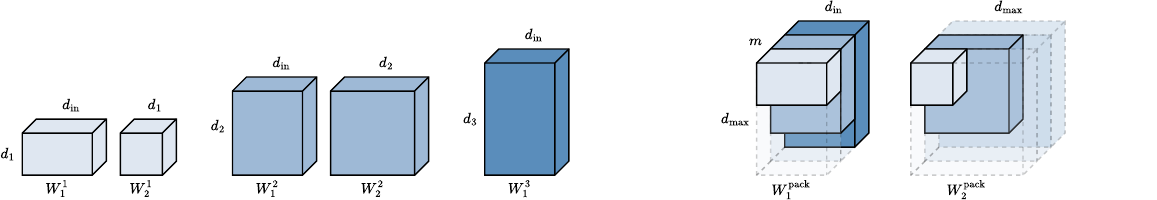}
    \caption{
        Packing multiple heterogeneous MLP backbones into one.
        \textit{(Left)}
        Weights of $m=3$ MLP backbones of different widths ($d_1 < d_2 < d_3$) and depths ($2$, $2$, $1$).
        Biases are omitted for simplicity.
        $W^i_j$ denotes the weights of the $j$-th layer in the $i$-th MLP.
        \textit{(Right)} Weights of the same $m=3$ MLPs as on the left, but now packed in one set of weights.
    }
    \label{fig:method-weights}
\end{figure*}

\begin{figure*}[h!]
    \centering
    \includegraphics[width=0.99\linewidth]{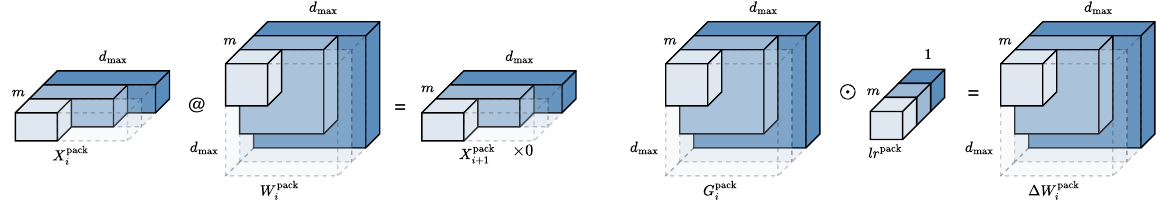}
    \caption{
        \textit{(Left)}
        A forward pass for a pack of linear layers.
        The input $X$ consists of $m$ object representations padded with zeros to $d_\text{max}$ and stacked along the pack dimension.
        The remaining notation follows \autoref{fig:method-weights}.
        First, a batched matrix multiplication applies, where all layers are applied in parallel to different input representations, as indicated by colors.
        Then, the output is masked according to the actual layer dimensions, as indicated by the ``$\times 0$'' label.
        \textit{(Right)}
        The weight-update computation for a pack of SGD optimizers, where $G$ denotes gradients.
        In practice, \method\ uses more advanced optimizers.
    }
    \label{fig:method-forward-backward}
\end{figure*}

As a model, \method\ represents $m$ MLPs, potentially with different hyperparameters, applied in parallel to $m$ inputs in a single forward pass.
We refer to this pattern as a \textit{model pack}, and to the individual models as \textit{base models}.
In this work, MLP is defined as a sequence of Linear--ReLU--Dropout blocks, so the varying MLP hyperparameters include depth, width, and dropout rates.
In principle, one can extend \method\ with residual connections \citep{he2016deep}, normalizations \citep{ioffe2015batch,ba2016layer}, and other custom layers, and vary base models by hyperparameters and presence or absence of these modules.

\textbf{Implementation.}
To pack heterogeneous base MLPs into one, we increase the width and depth of each base MLP to the maximum values across the pack and stack their parameters across a new pack dimension, as illustrated in \autoref{fig:method-weights}.
This allows applying all $m$ MLPs to $m$ input batches in parallel using batched matrix multiplications (e.g., \texttt{torch.bmm} in PyTorch \citep{paszke2019pytorch}).
To recover the behavior of the original base MLPs, we do the following:
\begin{itemize}[nosep,leftmargin=1em]
    \item
    For base models narrower than the maximum width, the corresponding parts of intermediate representations are zeroed out as shown on the left side of \autoref{fig:method-forward-backward}.

    \item
    For base models shallower than the maximum depth, only the necessary number of leading linear layers are applied, and the remaining deeper layers are skipped.

    \item Varying dropout rates are implemented by sampling $m$ dropout masks with $m$ different rates.
\end{itemize}

The described approach can be efficiently implemented in all mainstream DL frameworks using appropriate vectorized operations.
We note that to realize the ensemble potential of \method, one can usually use smaller maximum widths and depths than in traditional single models.
This, together with vectorization, alleviates the overhead from performing all matrix multiplications in $\mathbb{R}^{d_\text{max}}$ instead of $\mathbb{R}^{d_i}$ ($d_i \le d_\text{max}$) and from shallower models ``waiting'' for deeper models during the forward pass.

\subsection{Optimizer Pack}
\label{sec:method-optimizer-pack}

To support diverse training-related hyperparameters across base models, we implement \textit{optimizer packs}, analogously to model packs.
In optimizer packs, base optimizers differ in their learning rates, weight decays, and potentially other hyperparameters, such as momentum coefficients.
The right side of \autoref{fig:method-forward-backward} shows this mechanism for the vanilla SGD; in practice, we use more advanced optimizers, as discussed in \autoref{sec:method-practical-notes}.
For an optimizer pack, the state consists of individual optimizer states stacked along a new pack dimension, where tensor shapes in each individual state follow those of the deepest and widest base MLP.

\subsection{Ensemble}
\label{sec:method-ensemble}

Including all base models of \method\ in the final ensemble is not viable, since random hyperparameter sampling for base models (discussed in \autoref{sec:method-hyperparameters}) inevitably makes some base models poorly suited as ensemble members.
Thus, the ensemble must be constructed from base models selectively.
We consider two approaches for this: \textit{offline} and \textit{online}, both described later in this section.

\textbf{Greedy ensembling: a quick recap.}
Before presenting the offline and online approaches to ensembling, we briefly describe the greedy ensembling algorithm used by both as the core primitive for selecting ensemble members from a given pool of base models, also called \textit{candidates}.
We define greedy ensembling as follows:

\begin{itemize}[nosep,leftmargin=1em]
    \item
    Start by selecting the best base model according to the \textit{ensemble score} on the validation set.
    Unless stated otherwise, we use the target task metric as the ensemble score.
    In practice, depending on the dataset, using the training loss as the ensemble score is sometimes a better choice.

    \item
    In a loop, on each iteration, add to the ensemble the base model from the pool that improves the ensemble performance the most on the validation set (the individual model performance is ignored).
    Base models from the pool are added to the ensemble without replacement.

    \item
    Stop when the specified ensemble size $m_\text{ens}$ is reached or when no further improvement is possible.

\end{itemize}

While more advanced ensembling algorithms exist (see \citet{caruana2004ensemble} as an entry point to the topic), we found the greedy algorithm to be a decent simple baseline.

\newpage
\textbf{Offline ensemble.}
The simple \textit{offline} approach is to first train all base models independently (using \method\ as ``infrastructure'' for faster training) and then select the ensemble members.
Modulo \method's efficient implementation, this is the ``Tuned+Ensembled'' strategy used in \citet{ericson2025tabarena}.

The key strength of the offline approach is its simplicity.
At the same time, it has two important limitations.
First, intermediate base-model states remain unused, significantly limiting the candidate pool for ensemble construction, which can result in worse performance.
Second, there is no way to stop the base-model training based on ensemble performance, which again can hurt performance and also makes the total runtime equal to that of the slowest-to-train base model.

\textbf{Online ensemble.}
With the above in mind, \method\ uses the \textit{online} approach by default.
Generally, this means updating the ensemble continuously during training using intermediate states of the base models.
In this work, we rebuild the ensemble from scratch once per epoch.
At a given epoch, the pool of candidates consists of the latest base model states and the ensemble members from the previous epoch.
Because the previous ensemble members are added to the pool of candidates, multiple instances of the same base model from different epochs can become ensemble members, since such instances are treated as independent candidates.

\subsection{Training}
\label{sec:method-training}

\textbf{Base models.}
The base models of \method\ are trained in parallel on different batch sequences.
Thus, one training batch for \method\ consists of $m$ base batches of size $B$ stacked along the new pack dimension.
The models are trained independently: their \textit{losses} are aggregated, not their predictions.

\textbf{Early stopping for the ensemble.}
We stop training after $p_{\text{ens}} + 1$ consecutive epochs without improvement in ensemble performance on the validation set.

\textbf{Early stopping for the base models.}
To further accelerate \method training, we apply early stopping to base models as well, so that more compute is allocated to the still-improving base models.
Specifically, if a base model does not improve its validation performance for $p_{\text{base}} + 1$ consecutive epochs, it is removed from the pack.

\subsection{Hyperparameters}
\label{sec:method-hyperparameters}

Hyperparameters for the base models of \method\ are sampled randomly from user-defined hyperparameter spaces.
Thus, hyperparameters of these spaces (e.g., range bounds, distributions) are a part of \method's hyperparameters.
Other hyperparameters of \method\ include the number of base models $m$, early-stopping patience $p_{\text{base}}$ for base models and $p_{\text{ens}}$ for the ensemble, standard training-related options (e.g., a batch size), and ensembling-related options (e.g., the 
maximum ensemble size $m_\text{ens}$).

For simplicity, in this work we never tune \method's hyperparameters and always use $m=\nbasemodels$, $p_\text{base} = 16$, $p_\text{ens} = 32$, $m_\text{ens} = \maxensemblesize$ (to match the TabM baseline \citep{gorishniy2025tabm}), and the base hyperparameter spaces specified in \autoref{A:sec:impl-method}.
In particular, for simplicity, we use a fixed base-model width instead of sampling diverse values.
Overall, every performance number reported for \method\ is a result of a single run (modulo multi-seed runs for evaluation purposes).

\subsection{Practical Notes}
\label{sec:method-practical-notes}

\textbf{Base model family.}
We observe that the base model family has a significant impact on the ensemble performance, and techniques aimed at improving individual model performance remain highly relevant for the ensemble performance of \method.
In particular, we find that feature embeddings \citep{gorishniy2022embeddings} bring substantial benefits, and therefore recommend using them by default.
In this work, we use a variant of periodic embeddings (see \autoref{A:sec:feature-embeddings} for details), and diversify their hyperparameters similarly to those of the MLP backbone.

Overall, it seems that the ensembling mechanism behind \method\ efficiently realizes the potential of a given model family, but cannot fully compensate for the fundamental limitations of the base model family.
Thus, developing better base architectures remains a worthy research direction.

\textbf{Optimizer family.}
In line with \citet{gorishniy2026benchmarking}, we found Muon \citep{jordan2024muon} to consistently outperform AdamW \citep{loshchilov2019decoupled} both for \method\ and for prior baselines
(see \autoref{A:sec:optimizer-choice} for details).
Thus, we adopt Muon as the default optimizer for all methods in all experiments.

\textbf{Limitations} are discussed in \autoref{A:sec:limitations}.

\section{Experiments}
\label{sec:experiments}

In this section, we compare \method\ against tabular ML and DL baselines on public benchmarks.

\subsection{Models}
\label{sec:experiments-baselines}

We use the following baselines:
XGBoost \citep{chen2016xgboost} --- a powerful GBDT implementation; plain MLP, as in \citet{gorishniy2025tabm};
ModernNCA \citep{ye2024modern} --- a modern retrieval-based model; RealMLP \citep{holzmüller2024better} --- an advanced MLP-like architecture combined with a specific training recipe
(the recipe is unique to RealMLP and not used by \method\ by other baselines);
TabM \citep{gorishniy2025tabm} --- a parameter-efficient ensemble of MLPs;
TabICLv2 \citep{qu2026tabiclv2} and TabPFN-3 \citep{grinsztajn2026tabpfn3} --- modern tabular foundation models (TFMs).
See \autoref{A:sec:impl} for implementation details.

\textbf{\mlpens.}
Additionally, we introduce \mlpens\ (``HPE'' stands for \textbf{H}yper\textbf{P}arameter \textbf{E}nsemble) --- essentially a multi-process implementation of \method\ with offline ensembling, inspired by the ``Tuned+Ensembled'' evaluation regime in TabArena \citep{ericson2025tabarena}.
That is, all base models with randomly sampled hyperparameters are trained independently in $N_p$ concurrent processes on one GPU, and the ensemble is constructed afterwards.
In the experiments, we set $N_p$ to the value that reveals the full efficiency potential of \mlpens.
Although we expect \mlpens to lag behind \method\ in both effectiveness and efficiency, this baseline has its strengths.
In particular, it does not require reimplementing standard layers and optimizers, and it allows all ensemble-related logic to be concentrated in a separate base-model-agnostic pipeline.
Thus, we believe it is worth quantifying the difference between \method\ and its multi-process version to better understand the trade-offs.

\textbf{$\dagger$ Feature embeddings.}
Given the significant effect of feature embeddings \citep{gorishniy2022embeddings} on task performance, we explicitly mark models that use feature embeddings with $\dagger$: $\method^\dagger$, $\mathrm{MLP}^\dagger$, etc.
For a fair comparison, we use the same feature embeddings for all models, namely, a variant of periodic embeddings (see \autoref{A:sec:feature-embeddings}).

\begin{figure*}[h!]
    \centering
    \includegraphics[width=0.99\linewidth]{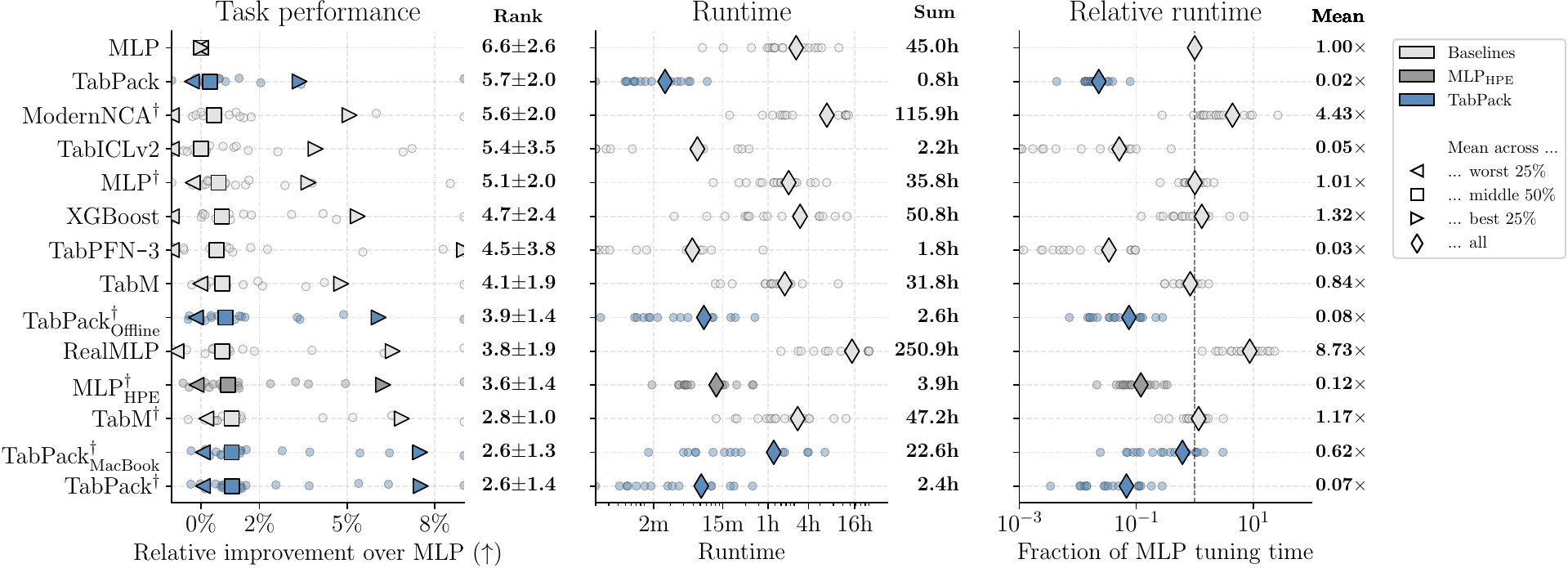}
    \caption{
        \textit{(Left)}
        Task performance of tabular models, on the \ndatasets\ datasets from \autoref{A:tab:dataset-properties}.
        The models are sorted by mean rank.
        Each dot represents the result on one dataset, and the triangle--square--triangle markers denote mean values across disjoint dataset groups (the groups vary between models) as explained in the legend.
        Triangle-related values are clipped to fit the plot.
        \textit{(Middle)}
        The absolute runtime on the same datasets as in the left plot.
        For non-foundation models, this is the hyperparameter tuning time.
        For foundation models, this is the inference time.
        For \method\ and \mlpens, this is the time of a single run, since there is no tuning.
        \textit{(Right)}
        Same as the middle plot, but all values are computed relative to the tuning time of MLP.
    }
    \label{fig:performance-runtime}
\end{figure*}

\newpage
\textbf{\method.}
We evaluate \method\ both without (\method) and with (\methodemb) feature embeddings.
Additional variants of \method include \methodoffline\ --- the version of \method\ with offline ensembling (see \autoref{sec:method-ensemble}), and \methodmacbook --- \method trained on a MacBook without a discrete GPU (see \autoref{A:sec:hardware}).

\subsection{Evaluation Protocols}
\label{sec:evaluation-protocol}

\textbf{Evaluation protocol for baselines.}
Most baselines undergo traditional hyperparameter tuning followed by multi-seed evaluation of the tuned configuration, as in \citet{gorishniy2025tabm}.
The only exceptions are foundation models, which are evaluated with their default hyperparameters and with the validation set merged into the training set.

\textbf{Evaluation protocol for \method}.
The somewhat unconventional workflow of \method\ requires a dedicated protocol for multi-seed evaluation.
There are two possible approaches:

\begin{itemize}[nosep,leftmargin=1em]
    \item
    (Optimistic)
    Run \method\ $N_\text{seeds}$ times with different random seeds and report performance and runtime aggregated over the seeds.

    \item
    (Conservative)
    First, perform the ``main'' run of \method\ once.
    Suppose that $M_\text{ens}$ base models were selected for the final ensemble; in this work, $M_\text{ens} \le m_\text{ens} = \maxensemblesize < \nbasemodels = m$.
    Then, rerun \method\ $N_\text{seeds}$ times with different random seeds using only the $M_\text{ens}$ selected base models.
    Report the task performance aggregated over the random seeds, and the runtime of the main run.
\end{itemize}

The conservative protocol naturally results in worse \method performance than the optimistic protocol, because its secondary runs use fewer base models.
Nevertheless, we use the \textit{conservative} protocol to better align with the tune-once routine of traditional models.

\textbf{Evaluation protocol for \mlpens.}
As with \method, we use the \textit{conservative} protocol for \mlpens, which means running \mlpens\ once, then rerunning it $N_{\text{seeds}}$ times with the subset of base models selected during the main run, and reporting aggregated task performance of the secondary runs and the runtime of the main run.

\subsection{Task Performance And Runtime}
\label{sec:experiments-performance}

\textbf{Results.}
We visualize the results in \autoref{fig:performance-runtime} and summarize them as follows:

\begin{itemize}[nosep,leftmargin=1em]
    \item
    \method\ shows strong out-of-the-box performance and substantial speedups over the traditional experiment workflow.
    Remarkably, \method enables faster experiment cycles on a modern MacBook than some prior methods on an enterprise-grade GPU, which is an NVIDIA A100 in our case.

    \item
    The online approach to ensembling is superior to the offline one (represented by \methodoffline\ and \mlpens), although the latter can still be of interest due to its simplicity, decent performance, and good efficiency.

    \item
    Performance-wise, \method is more stable than some of the baselines, as reflected by the ``Mean across worst 25\%'' performance markers on the left size of the figure.

    \item
    Additionally, in \autoref{sec:regression}, we discuss the preliminary observation that \method performs especially well on large regression tasks.

\end{itemize}

\newpage
\textbf{Main takeaway:} \method\ is a strong and efficient baseline for solving ML problems on tabular data, substantially reducing effort and compute needed to achieve competitive performance.

\subsection{Comparison across Time Budgets}
\label{sec:time-budgets}

We now compare the task performance of tabular models under varying time budgets.
Specifically, on a given dataset and for a given time budget, we measure the following:

\begin{itemize}[nosep,leftmargin=1em]
    \item
    For \method, the performance achieved in a single run when training is stopped after the given time budget.

    \item
    For \mlpens, the performance of the ensemble greedily built from the fully trained base models available after the given time budget.

    \item
    For traditional methods, the performance of the best hyperparameter configuration found when hyperparameter tuning is stopped after the given time budget.
    
\end{itemize}

\begin{figure}[h!]
    \centering
    \includegraphics[width=0.98\linewidth]{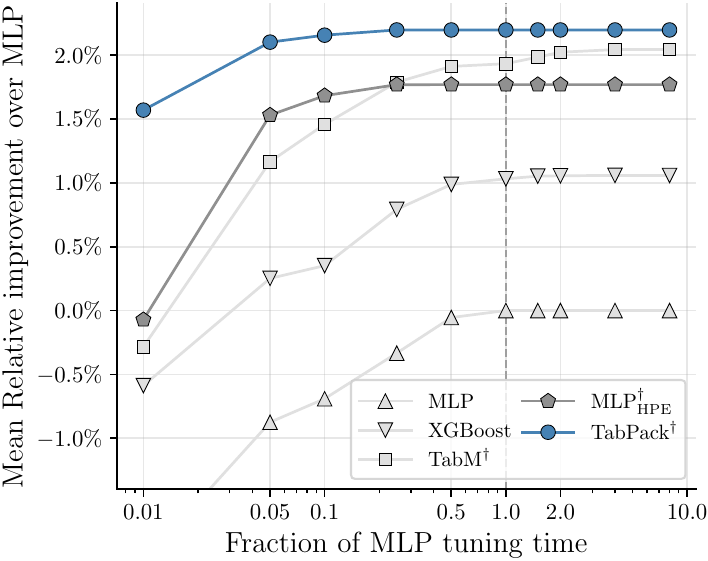}
    \caption{
        Performance of tabular models as a function of the time budget, averaged over the \ndatasets\ datasets from \autoref{A:tab:dataset-properties} and over three runs with different random seeds.
        See \autoref{sec:time-budgets} for details.
    }
    \label{fig:score-vs-time}
\end{figure}

The results in \autoref{fig:score-vs-time} indicate that \method\ is a strong model across a wide range of time budgets, further highlighting its potential as a go-to solution for practice and fast research experiments.
In particular, we note how quickly \method\ realizes most of its potential.
Additionally, we again highlight \mlpens\ as a decent baseline.

\textbf{Main takeaway:} \method\ is an effective tabular DL solution across various time budgets.

\subsection{Task Performance on Other Benchmarks}
\label{sec:performance-on-other-benchmarks}

After the joint evaluation of performance and runtime in the previous sections, we additionally evaluate task performance alone on other benchmarks, where full-fledged efficiency measurements are difficult and/or expensive to obtain.
We consider two additional benchmarks:
\begin{itemize}[nosep,leftmargin=1em]
    \item
    \textbf{Large datasets with 1M+ training objects}, where \method outperforms baselines on four out of five datasets.
    See \autoref{A:sec:evaluation-on-large-datasets}.

    \item
    \textbf{Small-to-medium datasets with IID splits from TabArena} \citep{ericson2025tabarena}, a benchmark that is more challenging for \method\ due to its smaller scale, where \method performs competitively overall, especially in the tuning-free regime.
    See \autoref{a:sec:evaluation-on-tabarena}.
\end{itemize}

\section{Analysis}
\label{sec:analysis}

\subsection{The Role of Hyperparameter Diversity}
\label{sec:hyperparameter-diversity}

Random hyperparameter sampling enables the tuning-free workflow of \method.
However, beyond its efficiency and convenience benefits, it is unclear whether diverse hyperparameters are actually important for task performance.
To this end, we additionally evaluate $\text{\method}_\text{SameHP}$: \method\ with all base models having the same hyperparameters tuned in a traditional manner on each dataset separately (see \autoref{A:sec:impl-tuning} for implementation details).
By analogy, the workflow $\text{\method}_\text{SameHP}$ is the same as that of TabM, where all base models also share the same hyperparameters, which must be tuned.
The results in \autoref{tab:hyperparameter-diversity} show that the tuning slightly improves the performance at the cost of noticeable increase of the total runtime.
Thus, random hyperparameter sampling is a reasonable default strategy; its main applied role is reducing the need for tuning rather than improving task performance by diversifying the base models.

\begin{table}[h!]
\centering
\caption{
    The performance and runtime on the same datasets as in \autoref{fig:performance-runtime} of three \method variants described in \autoref{sec:hyperparameter-diversity}.
    For $\text{\method}^\dagger_\text{SameHP}$, the runtime is the hyperparameter tuning time across all datasets, while for the rest it is the time of a single run.
}
\begin{tabular}{lcr}
\toprule
Model & Rank & Runtime \\
\midrule
\methodemb & $1.6 \pm 0.6$ & $2.4$h \\
$\text{\method}^\dagger_\text{DiverseWidths}$ & $1.6 \pm 0.5$ & $3.5$h \\
$\text{\method}^\dagger_\text{{SameHP}}$ & $1.4 \pm 0.7$ & $65.5$h \\
\bottomrule
\end{tabular}

\label{tab:hyperparameter-diversity}
\end{table}

A related question is how our choice to use the same width for all base models affects performance.
To answer this, we test \method variant where the base backbone width is sampled from $\mathrm{Uniform}[64,512]$, and we denote this variant as $\text{\method}_\text{DiverseWidths}$.
As shown in \autoref{tab:hyperparameter-diversity}, sampling varying model widths results in comparable performance with an increase in runtime caused by the higher maximum model width.
However, it is possible that the constant-width approach is competitive due to other details of our setup that implicitly compensate for the lack of diversity in model widths.
We also note that varying the width requires patching public Muon implementations to work correctly.

\subsection{A Closer Look at the Final Ensembles}
\label{sec:analysis-ensembles}

Now, we take a closer look at the ensembles produced by \method.
We will analyze the runs of \methodemb\ reported in \autoref{fig:performance-runtime}.

\textbf{Ensemble size.}
Across all datasets, the mean number of base models in the final ensemble is $16 \pm 9$ (as a reminder, the maximum allowed ensemble size is $m_\text{ens}=\maxensemblesize$).
For comparison, the ensemble size of TabM used in the original paper is $32$.
Thus, while \method\ uses more base models during its single run, the final ensemble sizes are comparable to those of prior work.
This, in particular, ensures practical inference efficiency, as shown in \autoref{A:sec:inference-efficiency}.

\textbf{Hyperparameter space utilization.}
Recall that the hyperparameters for base models of \method\ are sampled randomly from a predefined space.
To estimate to what extent this space is utilized by the final ensemble on a given dataset, for each hyperparameter, we divide its ensemble range (the difference between the maximum and the minimum values in the ensemble) by the total allowed range, and report this metric in \autoref{tab:hp-coverage}.
The results indicate a non-trivial hyperparameter diversity between the ensemble members.

\begin{table}[h!]
\centering
\caption{
    Hyperparameter space utilization, as defined in \autoref{sec:analysis-ensembles}, aggregated over the runs of \methodemb\ from \autoref{fig:performance-runtime}.
    $d_\text{emb}$ and $\sigma$ are feature embeddings hyperparameters (see \autoref{A:sec:feature-embeddings}).
}
\label{tab:hp-coverage}
\begin{tabular}{l l}
\toprule
\textbf{Hyperparameter} & \textbf{Space utilization} \\
\midrule
Depth & $0.88 \pm 0.27$ \\
Dropout & $0.77 \pm 0.28$ \\
$d_\text{emb}$ & $0.90 \pm 0.16$ \\
$\sigma$ & $0.81 \pm 0.19$ \\
Learning rate (Muon) & $0.72 \pm 0.24$ \\
Learning rate (AdamW) & $0.75 \pm 0.26$ \\
Weight decay & $0.68 \pm 0.16$ \\
\bottomrule
\end{tabular}

\end{table}

\subsection{Hyperparameter Exploration}
\label{sec:analysis-mdn}

\begin{figure}[h!]
    \centering
    \includegraphics[width=0.97\linewidth]{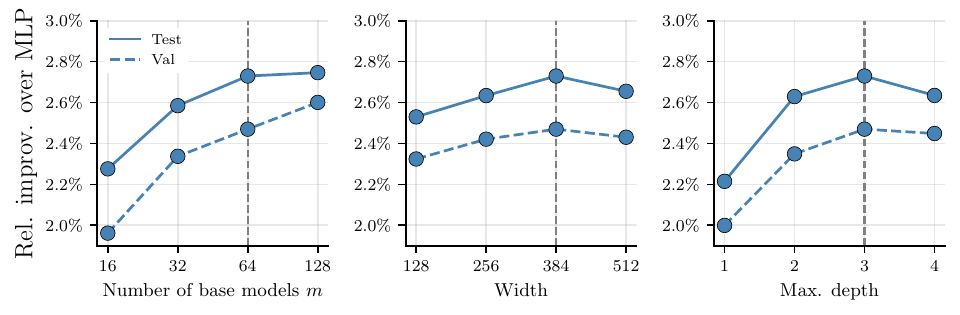}
    \caption{
        Task performance of \methodemb\ depending on different hyperparameters, averaged over the \ndatasets\ datasets from \autoref{A:tab:dataset-properties}.
        The vertical dashed lines correspond to the values used throughout the paper.
    }
    \label{fig:hyperparameter-exploration}
\end{figure}

Due to the limited presence of \method-like ensemble-first methods in the tabular DL landscape, it is unclear how the key model-related hyperparameters affect the task performance of the final ensemble.
To provide some intuition on this aspect, we vary the main architectural hyperparameters of \method\ and report the results in \autoref{fig:hyperparameter-exploration}.
From the results, we conclude that the configuration used in this work can serve as a starting point for future work.
However, we note that our results reflect \method's behavior on specific datasets and under specific experiment protocol.
For example, richer base hyperparameter spaces may require higher values of $m$ to fully realize \method's potential.

\section{Conclusion \& Future Work}
\label{sec:conclusion}

In this work, we presented \method --- an efficient ensemble of MLPs for solving machine learning problems on tabular data.
The key feature of \method is its ability to efficiently produce powerful ensembles with little to no tuning, substantially reducing effort and compute resources to achieve strong results and enabling significantly faster experimental cycles.
This makes \method an appealing baseline for practitioners and researchers.

Examples of  potential directions for future work include advancing the online ensemble algorithm behind \method, reducing the reliance of \method\ on the validation set, developing better base architectures for their use in \method.
Additionally, we separately highlight the opportunity to extend the \method-style hyperparameter diversification to other parts of the training pipeline, such as loss functions and data preprocessing, as well as the opportunity to vary base models not only by hyperparameters but also by architectural elements.

\section*{Impact Statement}

This paper presents work whose goal is to advance the field of Machine
Learning. There are many potential societal consequences of our work, none
which we feel must be specifically highlighted here.

\bibliography{references}
\bibliographystyle{icml2026}

\newpage
\appendix
\onecolumn

\section{Additional Analysis and Results}

\subsection{Evaluation on Datasets  with $\ge1$M training objects}
\label{A:sec:evaluation-on-large-datasets}

In this section, we compare tabular models on datasets with more than one million training objects.
Specifically, we use five industrial datasets with temporal splits summarized in \autoref{A:tab:big-dataset-properties}.
Due to the long runtimes, for the baselines, we reuse the hyperparameters tuned in the main experiments in \autoref{fig:performance-runtime} on subsampled versions of the same datasets.
For \method, we use the single-run tuning-free workflow as in the main text with the base model patience reduced to $p_{\text{base}} = 4$ and the ensemble patience reduced to $p_{\text{ens}} = 8$.
The results in \autoref{A:tab:big-datasets} indicate that \method successfully scales to large datasets outperforming baselines on four out of five tasks.

\begin{table}[h!]
\centering
\caption{
    The task performance on five large datasets from the TabReD benchmark \cite{rubachev2025tabred}.
    The metrics are ROC-AUC for Homecredit Default and RMSE for the rest of the datasets, averaged over three random seeds.
}

\label{A:tab:big-datasets}
\setlength{\tabcolsep}{4pt} 
\fontsize{9}{10}\selectfont 
\begin{tabular}{lcc|ccccc}
\toprule
 & Training set size & \#Features & $\mathrm{TabPack^\dagger}$ & $\mathrm{TabM^\dagger}$ & $\mathrm{MLP^\dagger}$ & XGBoost & $\mathrm{TabPFN\texttt{-}3}$ \\
Dataset &  &  &  &  &  &  &  \\
\midrule
Homecredit default & 1.1M & 696 & $0.8693$ & $\mathbf{0.8737}$ & $0.8689$ & $0.8728$ & 0.8494 \\
Delivery eta & 11.2M & 223 & $\mathbf{0.5404}$ & $0.5410$ & $0.5422$ & $0.5413$ & OOM \\
Weather & 13.6M & 103 & $\mathbf{1.3899}$ & $1.4065$ & $1.4567$ & $1.4158$ & OOM \\
Maps routing & 6.4M & 986 & $\mathbf{0.1572}$ & $0.1595$ & $0.1586$ & $0.1602$ & OOM \\
Cooking time & 9.1M & 192 & $\mathbf{0.4758}$ & $0.4780$ & $0.4780$ & $0.4782$ & OOM \\
\bottomrule
\end{tabular}

\end{table}

\subsection{Evaluation on TabArena}
\label{a:sec:evaluation-on-tabarena}

\begin{figure}[h!]
    \centering
    \includegraphics[width=0.95\linewidth]{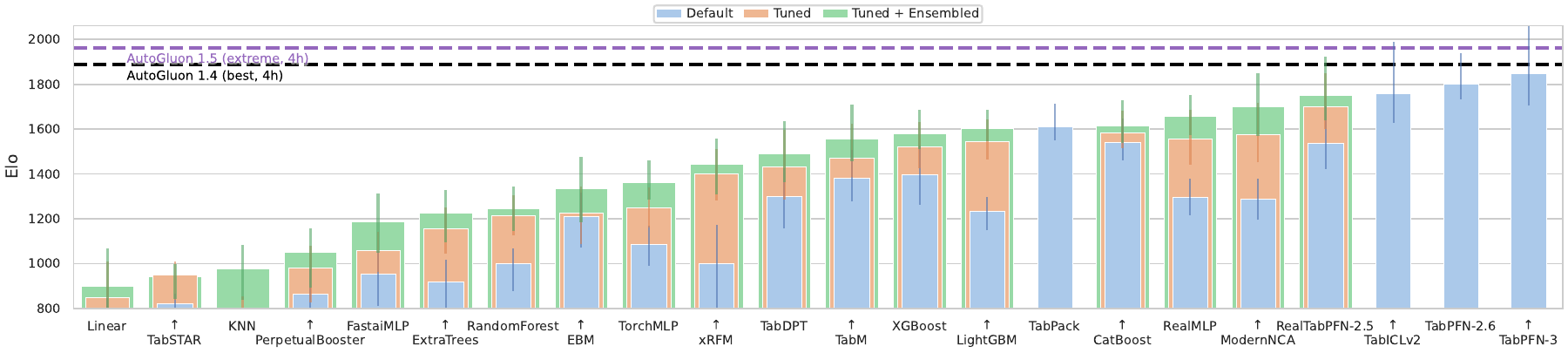}
    \caption{
        Results on \textit{medium-sized} datasets (at least 10K objects in total) from the TabArena-Lite benchmark \cite{ericson2025tabarena}.
    }
    \label{A:fig:tabarena-medium}
\end{figure}

\begin{figure}[h!]
    \centering
    \includegraphics[width=0.95\linewidth]{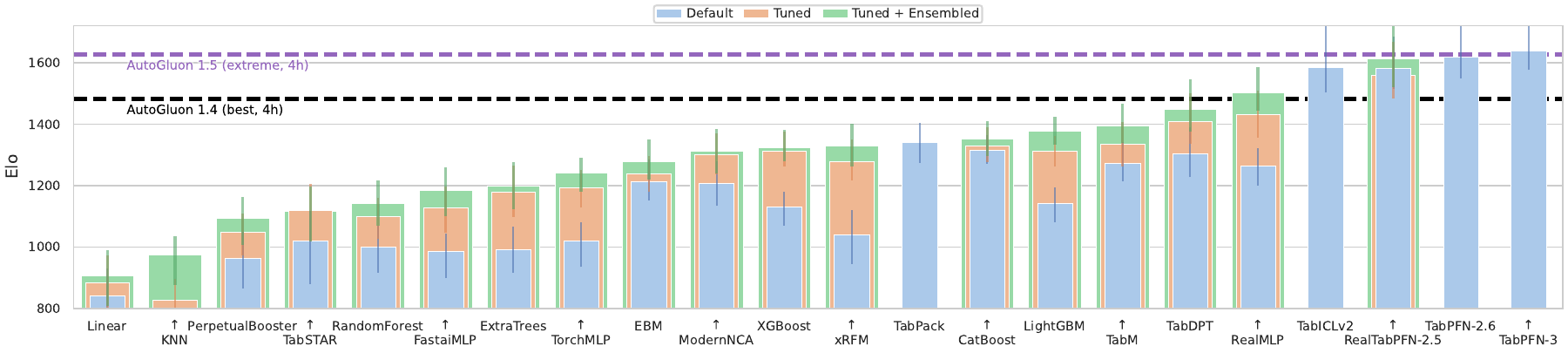}
    \caption{
        Results on \textit{small-sized} datasets (less than 10K objects in total) from the TabArena-Lite benchmark \cite{ericson2025tabarena}.
    }
    \label{A:fig:tabarena-small}
\end{figure}

In this section, we compare tabular models on the TabArena benchmark \citep{ericson2025tabarena} --- a benchmark consisting of of small-to-medium-sized datasets with IID random splits.
This may be a more challenging setup for \method, since TabArena datasets have validation sets of smaller absolute and relative sizes than those in our benchmark, and, as explained in \autoref{A:sec:limitations}, this may be problematic for \method.
Nevertheless, as shown in \autoref{A:fig:tabarena-medium} and \autoref{A:fig:tabarena-small}, \method demonstrates competitive performance on TabArena.
In particular, in the ``Default'' regime (the blue bars), \method confidently outperforms all models except foundation models, which are known to be strong on TabArena-like benchmarks..
\\\textbf{Implementation note:} in this section, we use the training loss as the ensemble score for the greedy ensembling algorithm powering \method, as we observed this variant to perform better on TabArena classification tasks.

\subsection{A Note on (Large) Regression Tasks}
\label{sec:regression}

A closer look at the results in \autoref{A:sec:evaluation-on-large-datasets} and \autoref{A:tab:per-dataset-results} reveals an interesting pattern: \method is not just competitive, but actually noticeably superior to competitors specifically on large regression tasks.
Due to the limited number of such datasets, it is hard to tell whether this is a general phenomenon or a coincidence.
Plus, specifically in \autoref{A:sec:evaluation-on-large-datasets}, the baselines did not undergo the full-fledged hyperparameter tuning due to the high cost of experiments.

As a thought experiment, let's assume that the strong performance of \method on large regression tasks is a general pattern.
Then, the underlying reason for this pattern may be the combination of the following factors:
\begin{itemize}[nosep, leftmargin=2em]
    \item
    By its nature, the greedy ensembling algorithm used by \method gradually refines the ensemble prediction.

    \item 
    The online approach to ensembling used by \method greatly expands the set of ensemble member candidates considered by \method, which further strengthens \method's ability to gradually improve the ensemble during training.

    \item
    Unlike more ``discrete'' classification metrics, such as accuracy, typical regression metrics, such as RMSE or $R^2$, are more rewarding for gradual prediction improvements.
    In turn, using more ``continuous'' metrics, such as cross-entropy, to guide ensemble construction on classification tasks creates the discrepancy between the ensemble score and the actual task metric (although it can be beneficial sometimes; for example, see \autoref{a:sec:evaluation-on-tabarena}).
    Regression metrics such as RMSE or $R^2$ are free from this problem.

    \item
    On average, the larger the dataset, the better its validation set serves as a proxy for its test set.
    This helps \method to generalize well to the test set despite using the validation set to guide ensemble construction.
    
\end{itemize}

\subsection{Muon vs. AdamW}
\label{A:sec:optimizer-choice}

As mentioned in the main text, we find Muon \citep{jordan2024muon} to consistently outperform AdamW \citep{loshchilov2019decoupled} as the optimizer for \method.
Specifically, when compared under exactly the same experiment protocol as in \autoref{sec:experiments-performance} across all \ndatasets datasets from \autoref{A:tab:dataset-properties}:
\begin{itemize}[nosep,leftmargin=2em]
    \item
    For \method, the win/tie/loss counts of Muon against AdamW are 9/3/5.

    \item
    For \methodemb, the win/tie/loss counts of Muon against AdamW are 13/3/1.
\end{itemize}

Where ``win'' and ``loss'' simply mean ranks $1$ and $2$, respectively, and ``tie'' means that both methods have rank $1$.

\subsection{Inference Efficiency}
\label{A:sec:inference-efficiency}

In this section, we quickly estimate how \method compares to prior work in terms of inference efficiency.
We start with two observations:
\begin{itemize}[nosep,leftmargin=2em]
    \item
    The TabM paper \citep{gorishniy2025tabm} contains detailed measurements of the inference throughput of tabular models on GPU and CPU, showing that TabM exhibits practical inference characteristics suitable for many real-world use cases.

    \item
    In this paper, the ensembles produced by \method are almost surely more inference-efficient than TabM, because: (1) as mentioned in \autoref{sec:analysis-ensembles}, the final ensemble size of \method is smaller than that of TabM; (2) the base model width used in \method is \defaultwidth, while in TabM it is tuned in $\mathrm{Uniform}[64,1024]$; (3) the maximum base model depth of \method is \defaultdepth, while in TabM it is tuned in $\mathrm{Uniform}[1,5]$.
\end{itemize}

\begin{figure}[h!]
\centering
\includegraphics[width=0.5\textwidth]{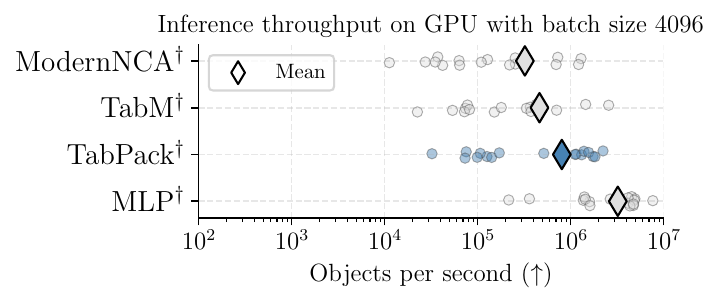}
\caption{
    Inference throughput on NVIDIA A100 on \ndatasets datasets from \autoref{A:tab:dataset-properties}
}
\label{A:fig:inference-throughput}
\end{figure}

With the above in mind, it is reasonable to expect that \method is at least as efficient as TabM.
To validate this intuition, we perform a quick experiment and report the results in \autoref{A:fig:inference-throughput}.
The results are in line with the expectations and establish \method as a practical solution for real-world usage.

\subsection{Memory Consumption during Training}
\label{A:sec:memory-consumption}

In \autoref{A:tab:memory-consumption}, we report the memory consumption of \method during training.
The results indicate that \method training will often fit in a consumer-grade GPU.
That said, for large enough number of base models and large enough number of features in a dataset, \method may require hardware with more than 12GB memory to run.

\begin{table}[h!]
\centering
\caption{
    Peak memory usage in GB during training averaged over \ndatasets datasets from \autoref{A:tab:dataset-properties} depending on the number of base models.
}
\setlength{\tabcolsep}{4pt}
\fontsize{10}{10}\selectfont
\begin{tabular}{lcrrrrrrrr}
\toprule
Num base models: & $1$ & $2$ & $4$ & $8$ & $16$ & $32$ & $64$ & $128$ \\
\midrule
$\mathrm{TabM^\dagger}$ & 0.3 & 0.4 & 0.5 & 0.7 & 1.2 & 2.1 & 4.0 & 7.7 \\
$\methodemb$ & 0.3 & 0.4 & 0.6 & 0.9 & 1.6 & 3.0 & 5.8 & 11.3 \\
\bottomrule
\end{tabular}

\label{A:tab:memory-consumption}
\end{table}

\section{Feature Embeddings}
\label{A:sec:feature-embeddings}

\textbf{Background.}
An embedding for continuous features \citep{gorishniy2022embeddings} is a mapping $f_\text{emb}: \mathbb{R} \rightarrow \mathbb{R}^{d_\text{emb}}$ applied to each continuous feature in isolation \textit{before} mixing the features in the main backbone, where $d_\text{emb}$ is the embedding size.
A simple example of a non-linear feature embedding is $f_\text{emb}(x_i) = \mathrm{ReLU(w_i \cdot x + b_i)}$, where $i$ is the feature index, and $w_i, b_i \in \mathbb{R}^d_\text{emb}$ are trainable parameters (not shared between features).
When feature embeddings are used with MLP-like models, the backbone takes the input consisting of all feature embeddings concatenated in one flat vector.

\textbf{New periodic embeddings.}
In this work, we introduce a new variation of ``periodic embeddings'', i.e. embeddings using periodic activation functions under the hood.
Prior work includes PLR embeddings from \citet{gorishniy2022embeddings} and PBLD embeddings from \citet{holzmüller2024better}.
Our variant adopts certain design elements from prior work and improves efficiency noticeably by avoiding hidden layers.

Our periodic embedding $f_\text{periodic}: \mathbb{R} \rightarrow \mathbb{R}^{d_\text{emb}}$ is formally defined as follows:
$$
    f_\text{periodic}(x_i) = \alpha_i \cdot [x_i, \cos(w_i \cdot x + b_i)] + \beta_i
$$
where $\cdot$ denotes element-wise multiplication, $+$ denotes element-wise addition, $[\ldots]$ denotes concatenation, $i$ is the feature index, $w_i \in \mathbb{R}^{d_\text{emb} - 1}$ is initialized from $\mathrm{N}(0,\sigma)$ ($\sigma \in \mathbb{R}$ is a hyperparameter), $b_i  \in \mathbb{R}^{d_\text{emb} - 1}$ is initialized with zeros, and $\alpha_i,\beta_i$ are also initialized with zeros.

\textbf{Notation.}
In this work, all models marked with $\dagger$ use our periodic embeddings described above.

\section{Limitations}
\label{A:sec:limitations}

We highlight the following limitations of \method:

\begin{itemize}[nosep,leftmargin=1em]
    \item
    In our current implementation, \method\ relies on the validation set for both ensemble construction and early stopping.
    We hypothesize that overfitting to the validation set may become an issue if the validation set is small or there is a significant distribution shift between the validation and test sets.

    \item
    \method\ naturally requires more GPU memory for training compared to traditional single models.
    We report memory usage in \autoref{A:sec:memory-consumption}.
    In GPU-memory-constrained setups, one can reduce the maximum base model dimensions (in particular, feature embedding sizes) and/or the number of base models to alleviate the issue.
    Another strategy is a fallback to offline ensembles: run \method\ multiple times with the reduced number of base models, and build an ensemble using all base models afterwards.

    \item 
    To extend \method\ with new architectural elements and optimizers, one has to implement their packed versions instead of simply reusing existing implementations.
    We hope that our code base will serve as a helpful example for implementing new packed building blocks.
\end{itemize}

\section{Hardware}
\label{A:sec:hardware}

All experiments, modulo the runs of \methodmacbook, were conducted in the single-GPU regime using NVIDIA A100 80GB.

For \methodmacbook, we used a 16-inch MacBook Pro with the Apple M4 Pro chip with 10 performance CPU cores, 4 efficiency CPU cores, 20 GPU cores and 48GB of unified RAM.
The compute backend was \texttt{torch.device("mps")}, so the integrated GPU was used as the main compute device.
Crucially, in the macOS ``Battery'' settings, ``Energy Mode'' was set to ``High Power'' to sustain maximum training speed.

\section{Datasets}
\label{A:sec:datasets}

\autoref{A:tab:dataset-properties} summarizes properties of the datasets used in most experiments throughout the paper, while \autoref{A:tab:big-dataset-properties} summarizes properties of the datasets used specifically in \autoref{A:sec:evaluation-on-large-datasets}.
Note that some of the TabReD datasets are presented in both tables.
This is because they have two official versions, full and the subsampled, both introduced in the TabReD paper.
For \autoref{A:sec:evaluation-on-large-datasets}, we prepare variants of these datasets with full training sets, but samsampled validation and test sets, to make the results comparable between the two versions.

\begin{table}[h]
\centering
\caption{
    Extended properties of datasets used in experiments throughput the paper.
    Here, ``\# Train'', ``\# Val'', ``\# Test'' denotes the size of the corresponding dataset split; similarly, ``\# Num'', ``\# Bin'', ``\# Cat'' denotes the number of numerical, binary, and categorical features, respectively.
}
\label{A:tab:dataset-properties}
\setlength{\tabcolsep}{4pt} 
\fontsize{9}{10}\selectfont 
\resizebox{\textwidth}{!}{
\begin{tabular}{lcccccccccc}
\toprule
Name & Source & \# Train & \# Val & \# Test & \# Num & \# Bin & \# Cat & Task Type & Batch Size\\
\midrule
Churn Modelling &  \multirow{8}{*}{\begin{tabular}[c]{@{}c@{}}\textsc{TabM}\\ {\citep{gorishniy2025tabm}}\end{tabular}}  & $6\,400$ & $1\,600$ & $2\,000$ & $7$ & $3$ & $1$ & Binclass & 128 \\
California Housing & & $13\,209$ & $3\,303$ & $4\,128$ & $8$ & $0$ & $0$ & Regression & 256 \\
House 16H & & $14\,581$ & $3\,646$ & $4\,557$ & $16$ & $0$ & $0$ & Regression & 256 \\
Adult & & $26\,048$ & $6\,513$ & $16\,281$ & $6$ & $1$ & $8$ & Binclass & 256 \\
Diamond & & $34\,521$ & $8\,631$ & $10\,788$ & $6$ & $0$ & $3$ & Regression & 512 \\
Otto Group Products & & $39\,601$ & $9\,901$ & $12\,376$ & $93$ & $0$ & $0$ & Multiclass & 512 \\
Higgs Small & & $62\,751$ & $15\,688$ & $19\,610$ & $28$ & $0$ & $0$ & Binclass & 512 \\
Black Friday & & $106\,764$ & $26\,692$ & $33\,365$ & $4$ & $1$ & $4$ & Regression & 512 \\
Microsoft & & $723\,412$ & $235\,259$ & $241\,521$ & $131$ & $5$ & $0$ & Regression & 1024 \\
\midrule
Sberbank Housing & \multirow{8}{*}{\begin{tabular}[c]{@{}c@{}}\textsc{TabReD}\\ {\citep{rubachev2025tabred}}\end{tabular}} & 18 847 & 4 827 & 4 647 & 365 & 17 & 10 & Regression & 256 \\
Ecom Offers & & $109\,341$ & $24\,261$ & $26\,455$ & $113$ & $6$ & $0$ & Binclass & 1024 \\
Maps Routing & & $160\,019$ & $59\,975$ & $59\,951$ & $984$ & $0$ & $2$ & Regression & 1024 \\
Homesite Insurance & & $224\,320$ & $20\,138$ & $16\,295$ & $253$ & $23$ & $23$ & Binclass & 1024 \\
Cooking Time & & $227\,087$ & $51\,251$ & $41\,648$ & $186$ & $3$ & $3$ & Regression & 1024 \\
Homecredit Default & & $267\,645$ & $58\,018$ & $56\,001$ & $612$ & $2$ & $82$ & Binclass & 1024 \\
Delivery ETA & & $279\,415$ & $34\,174$ & $36\,927$ & $221$ & $1$ & $1$ & Regression & 1024 \\
Weather & & $340\,596$ & $42\,359$ & $40\,840$ & $100$ & $3$ & $0$ & Regression & 1024 \\
\bottomrule
\end{tabular}}

\end{table}

\begin{table}[h]
\centering
\caption{
    Extended properties of the datasets used in \autoref{A:sec:evaluation-on-large-datasets}.
    See \autoref{A:sec:datasets} for details.
}
\label{A:tab:big-dataset-properties}
\setlength{\tabcolsep}{4pt} 
\fontsize{9}{10}\selectfont 
\begin{tabular}{lcccc}
\toprule
Name & Source & \# Train & Other properties\\
\midrule
Maps Routing & \multirow{5}{*}{\begin{tabular}[c]{@{}c@{}}\textsc{TabReD} \\ with full training sets \\ {\citep{rubachev2025tabred}}\end{tabular}} &  6\,408\,198 & \multirow{5}{*}{\begin{tabular}[c]{@{}c@{}}Same as in \autoref{A:tab:dataset-properties}\end{tabular}} \\
Cooking Time & & 9\,096\,168 &  \\
Homecredit Default & & 1\,070\,762 &  \\
Delivery ETA & & 11\,151\,785 &  \\
Weather & & 13\,625\,138 &  \\
\bottomrule
\end{tabular}

\end{table}

\section{Implementation details}
\label{A:sec:impl}

\subsection{Experiment Setup}
\label{A:sec:impl-experiment-setup}

We mostly rely on the experiment setup from \citet{gorishniy2025tabm}.
As such, a significant portion of the below text  is copied from that work, with some modifications related to our study.

\textbf{Data preprocessing.}
For each dataset, for all DL-based solutions, the same preprocessing was used for fair comparison.
For numeric features, by default, we used a modified version of the quantile normalization from the Scikit-learn package \citep{pedregosa2011scikit}, with rare exceptions when it turned out to be detrimental (for such datasets, we used the standard normalization or no normalization).
For categorical features, we used one-hot encoding.
Binary features (i.e. the ones that take only two distinct values) are treated as categorical features. For the datasets from TabReD\cite{rubachev2025tabred}, we follow the data preprocessing from the original paper.

\textbf{Training baseline neural networks.}
For DL-based algorithms, we minimize cross-entropy for classification problems and mean squared error for regression problems.
We use the Muon optimizer \citet{jordan2024muon} for the main (rectangular) weights of linear layers in the main backbones, and the AdamW optimizer \citep{loshchilov2019decoupled} for everything else.
We do not apply learning rate schedules.
We do not use data augmentations.
For each dataset, we used a predefined dataset-specific batch size provided in \autoref{A:tab:dataset-properties}.
We continue training until there are $\texttt{patience}$ + 1 consecutive epochs without improvements on the validation set; we set $\texttt{patience} = 16$ for DL models.

\textbf{Hyperparameter tuning.}
In most cases, hyperparameter tuning is performed with the TPE sampler (typically, 50-100 iterations) from the Optuna package \citep{akiba2019optuna}.
Hyperparameter tuning spaces for most models are provided in individual sections below.

\textbf{Evaluation.}
On a given dataset, for a given model, the tuned hyperparameters are evaluated under multiple (in most cases, $5$ or $10$ random seeds.
The mean test metric and its standard deviation over these random seeds are then used to compare algorithms.

\subsection{Implementation Details of \autoref{sec:hyperparameter-diversity}}
\label{A:sec:impl-tuning}

To obtain $\method_\text{SameHP}^\dagger$, we use the standard hyperparameter tuning pipeline as for DL baselines in this work, i.e. it the standard tuning process using the TPE sampler from Optuna \citep{akiba2019optuna}.
In this tuning process, the hyperparameter tuning space is exactly the same as the one used for random hyperparameter sampling in \method.

\subsection{\method}
\label{A:sec:impl-method}

The hyperparameter space used to sample base model hyperparameters in \methodemb\ is provided in 
\autoref{A:tab:method-emb-space}.
For \method, it is the same, but without feature embedding hyperparameters and the maximum allowed depth is 4 instead of \defaultdepth.

\begin{table}[h!]
\centering
\caption{The hyperparameter sampling space for \methodemb\  base models}
{\renewcommand{\arraystretch}{1.2}
\begin{tabular}{ll}
    \toprule
    Parameter           & Distribution \\
    \midrule
    \# layers           & $\mathrm{UniformInt}[1,3]$ \\
    Width (hidden size) & $384$ \\
    Dropout rate        & $\{0.0, \mathrm{Uniform}[0.0,0.5]\}$ \\
    $d_\text{emb}$ & $\mathrm{UniformInt}[8,32]$ (step 4) \\
    $\sigma$         & $\mathrm{LogUniform}[1e\text{-}2, 10]$ \\
    Muon learning rate  & $\mathrm{LogUniform}[1e\text{-}3, 1e\text{-}1]$ \\
    AdamW learning rate & $\mathrm{LogUniform}[1e\text{-}4, 5e\text{-}3]$ \\
    Weight decay        & $\mathrm{LogUniform}[1e\text{-}3, 1]$ \\
    \bottomrule
\end{tabular}}
\label{A:tab:method-emb-space}
\end{table}

\subsection{MLP}

All MLP variants use the Muon optimizer for main network parameters and AdamW for remaining parameters (e.g., embeddings, biases). We train all MLP models in float32 precision.
\autoref{A:tab:mlp-space} and \autoref{A:tab:mlp-emb-space} provide the hyperparameter tuning spaces for MLP and MLP$^\dagger$ with the new periodic embeddings, respectively.

\begin{table}[h!]
\centering
\caption{The hyperparameter tuning space for MLP.}
{\renewcommand{\arraystretch}{1.2}
\begin{tabular}{ll}
    \toprule
    Parameter           & Distribution \\
    \midrule
    \# layers           & $\mathrm{UniformInt}[1,6]$ \\
    Width (hidden size) & $\mathrm{UniformInt}[64,1024]$ \\
    Dropout rate        & $\{0.0, \mathrm{Uniform}[0.0,0.5]\}$ \\
    Muon learning rate  & $\mathrm{LogUniform}[1e\text{-}4, 3e\text{-}2]$ \\
    AdamW learning rate & $\mathrm{LogUniform}[3e\text{-}5, 1e\text{-}3]$ \\
    Weight decay        & $\mathrm{LogUniform}[1e\text{-}3, 1]$ \\
    \midrule
    \# Tuning iterations & 100 \\
    \bottomrule
\end{tabular}}
\label{A:tab:mlp-space}
\end{table}

\begin{table}[h!]
\centering
\caption{The hyperparameter tuning space for MLP with embeddings.}
{\renewcommand{\arraystretch}{1.2}
\begin{tabular}{ll}
    \toprule
    Parameter           & Distribution \\
    \midrule
    \# layers           & $\mathrm{UniformInt}[1,5]$ \\
    Width (hidden size) & $\mathrm{UniformInt}[64,1024]$ \\
    Dropout rate        & $\{0.0, \mathrm{Uniform}[0.0,0.5]\}$ \\
    $d_\text{emb}$ & $\mathrm{UniformInt}[8,32]$ (step 4) \\
    $\sigma$         & $\mathrm{LogUniform}[1e\text{-}2, 10]$ \\
    Muon learning rate  & $\mathrm{LogUniform}[1e\text{-}4, 3e\text{-}2]$ \\
    AdamW learning rate & $\mathrm{LogUniform}[3e\text{-}5, 1e\text{-}3]$ \\
    Weight decay        & $\mathrm{LogUniform}[1e\text{-}3, 1]$ \\
    \midrule
    \# Tuning iterations & 100 \\
    \bottomrule
\end{tabular}}
\label{A:tab:mlp-emb-space}
\end{table}

\subsection{TabM}

All TabM variants use the Muon optimizer for main network parameters and AdamW for the remaining parameters.
We use $k=32$ for all TabM experiments.
We train TabM baselines in BFloat16 precision.
\autoref{A:tab:tabm-space} and \autoref{A:tab:tabm-emb-space} provide the hyperparameter tuning spaces for TabM and TabM$^\dagger$ with the new embeddings, respectively.

\begin{table}[h!]
\centering
\caption{The hyperparameter tuning space for TabM. Here, (B) = \{Black Friday, Microsoft, and all TabReD datasets\}. (A) contains all other datasets.}
{\renewcommand{\arraystretch}{1.2}
\begin{tabular}{ll}
    \toprule
    Parameter           & Distribution or Value \\
    \midrule
    $k$                 & $32$ \\
    \# layers           & $\mathrm{UniformInt}[1,5]$ \\
    Width (hidden size) & $\mathrm{UniformInt}[64,1024]$ \\
    Dropout rate        & $\{0.0, \mathrm{Uniform}[0.0,0.5]\}$ \\
    Muon learning rate  & $\mathrm{LogUniform}[1e\text{-}3, 1e\text{-}1]$ \\
    AdamW learning rate & $\mathrm{LogUniform}[1e\text{-}4, 5e\text{-}3]$ \\
    Weight decay        & $\mathrm{LogUniform}[1e\text{-}3, 1]$ \\
    \midrule
    \# Tuning iterations & (A) 100 (B) 50 \\
    \bottomrule
\end{tabular}}
\label{A:tab:tabm-space}
\end{table}

\begin{table}[h!]
\centering
\caption{The hyperparameter tuning space for TabM with embeddings. Here, (B) = \{Black Friday, Microsoft, and all the TabReD datasets.\}(A) contains all other datasets.}
{\renewcommand{\arraystretch}{1.2}
\begin{tabular}{ll}
    \toprule
    Parameter           & Distribution or Value \\
    \midrule
    $k$                 & $32$ \\
    \# layers           & $\mathrm{UniformInt}[1,4]$ \\
    Width (hidden size) & $\mathrm{UniformInt}[64,1024]$ \\
    Dropout rate        & $\{0.0, \mathrm{Uniform}[0.0,0.5]\}$ \\
    $d_\text{emb}$ & $\mathrm{UniformInt}[8,32]$ (step 4) \\
    $\sigma$         & $\mathrm{LogUniform}[1e\text{-}2, 10]$ \\
    Muon learning rate  & $\mathrm{LogUniform}[1e\text{-}3, 1e\text{-}1]$ \\
    AdamW learning rate & $\mathrm{LogUniform}[1e\text{-}4, 5e\text{-}3]$ \\
    Weight decay        & $\mathrm{LogUniform}[1e\text{-}3, 1]$ \\
    \midrule
    \# Tuning iterations & (A) 100 (B) 50 \\
    \bottomrule
\end{tabular}}
\label{A:tab:tabm-emb-space}
\end{table}

\subsection{ModernNCA}

We adapted the official implementation from \citet{ye2024modern}, with modifications according to our experiment protocol. We use the new modified periodic embeddings and train in float32 precision.

\autoref{A:tab:modernnca-space} provides the hyperparameter tuning space for ModernNCA$^\dagger$. When tuning the ModernNCA model on larger datasets (this includes all TabReD datasets without Ecom Offers and Sberbank Housing and Microsoft) we set a timeout on the tuning procedure at 12 hours.

\begin{table}[h!]
\centering
\caption{The hyperparameter tuning space for ModernNCA$^\dagger$. Here, (B) = \{Microsoft and TabReD datasets except Ecom Offers and Sberbank Housing\}. (A) contains all other datasets.}
{\renewcommand{\arraystretch}{1.2}
\begin{tabular}{ll}
    \toprule
    Parameter           & Distribution \\
    \midrule
    \# blocks           & $\mathrm{UniformInt}[0, 2]$ \\
    dim                 & $\mathrm{UniformInt}[64,1024]$ \\
    $d_{\mathrm{block}}$ & $\mathrm{UniformInt}[64,1024]$ \\
    Dropout rate        & $\mathrm{Uniform}[0.0,0.5]$ \\
    Sample rate         & $\mathrm{Uniform}[0.05, 0.6]$ \\
    $d_\text{emb}$ & (A) $\mathrm{UniformInt}[8,32]$ (B) $\mathrm{UniformInt}[4,16]$ (step 4) \\
    $\sigma$         & $\mathrm{LogUniform}[1e\text{-}2, 10]$ \\
    Muon learning rate  & $\mathrm{LogUniform}[1e\text{-}4, 3e\text{-}2]$ \\
    AdamW learning rate & $\mathrm{LogUniform}[3e\text{-}5, 1e\text{-}3]$ \\
    Weight decay        & $\mathrm{LogUniform}[1e\text{-}3, 1]$ \\
    \midrule
    \# Tuning iterations & (A) 100 (B) 50 \\
    \bottomrule
\end{tabular}}
\label{A:tab:modernnca-space}
\end{table}

\subsection{RealMLP}

We used the \texttt{tabarena-new} preset from pytabkit\footnote{\url{https://github.com/dholzmueller/pytabkit/tree/c126ea5}} and TPE sampler instead of random one in the original implementation. We also adapted Muon optimizer for RealMLP and tuned Muon learning rate with $\mathrm{LogUniform}[0.02, 0.3]$ space.

\subsection{TabICL-V2}

We used the official \texttt{tabicl} Python package\footnote{https://github.com/soda-inria/tabicl}.
We merged the validation set to the training set.
We used ordinal encoding for categorical features.
All other hyperparameters were kept at their default values.

\subsection{TabPFN-3}

We used the official \texttt{tabpfn} Python package\footnote{https://github.com/PriorLabs/TabPFN}.
We merged the validation set to the training set.
We used ordinal encoding for categorical features and pass them using \texttt{categorical\_feature\_indices}.
All other hyperparameters were kept at their default values.

\subsection{XGBoost}

\autoref{A:tab:xgboost-space} provides the hyperparameter tuning space for XGBoost.

\begin{table}[h!]
\centering
\caption{The hyperparameter tuning space for XGBoost.}
{\renewcommand{\arraystretch}{1.2}
\begin{tabular}{ll}
    \toprule
    Parameter           & Distribution or Value \\
    \midrule
    n\_estimators       & $4000$ \\
    max\_depth          & $\mathrm{UniformInt}[3,14]$ \\
    learning\_rate      & $\mathrm{LogUniform}[1e\text{-}3, 1]$ \\
    gamma               & $\{0, \mathrm{LogUniform}[1e\text{-}3, 100]\}$ \\
    lambda              & $\{0, \mathrm{LogUniform}[0.1, 10]\}$ \\
    min\_child\_weight  & $\mathrm{LogUniform}[1e\text{-}4, 100]$ \\
    subsample           & $\mathrm{Uniform}[0.5, 1.0]$ \\
    colsample\_bytree   & $\mathrm{Uniform}[0.5, 1.0]$ \\
    early\_stopping\_rounds & $200$ \\
    \midrule
    \# Tuning iterations & 200 \\
    \bottomrule
\end{tabular}}
\label{A:tab:xgboost-space}
\end{table}

\section{Per-Dataset Results}
\label{A:sec:per-dataset-results}

The per-dataset results are reported in the tables below.

\newcommand{\topalign}[1]{%
\vtop{\vskip 0pt #1}}

\begin{longtable}{p{0.5\textwidth}p{0.5\textwidth}}
\caption{
    Per-dataset performance for the methods.
    For each dataset, we report the mean and standard deviation of test metric across ten random seeds.
}
\label{tab:appendix-all-results}\\

\topalign{
\centering
\setlength\tabcolsep{2.5pt}
\renewcommand{\arraystretch}{0.8}
\begin{tabular}{ll}

\multicolumn{2}{c}{\small{Churn \textuparrow}} \\
\toprule
{\small Method} & {\small Score} \\
\midrule\\[-0.7cm]
\multicolumn{2}{c}{} \\[0.05cm]
{\footnotesize $\mathrm{MLP}$ } & {\footnotesize$0.8562 \pm 0.0018$}\\ 
{\footnotesize $\mathrm{XGBoost}$ } & {\footnotesize$0.8605 \pm 0.0020$}\\ 
{\footnotesize $\mathrm{TabICLv2}$ } & {\footnotesize$0.8658 \pm 0.0008$}\\ 
{\footnotesize $\mathrm{TabPFN\texttt{-}3}$ } & {\footnotesize$0.8655 \pm 0.0009$}\\ 
{\footnotesize $\mathrm{ModernNCA^\dagger}$ } & {\footnotesize$0.8591 \pm 0.0027$}\\ 
{\footnotesize $\mathrm{RealMLP}$ } & {\footnotesize$0.8599 \pm 0.0023$}\\ 
{\footnotesize $\mathrm{TabM}$ } & {\footnotesize$0.8638 \pm 0.0017$}\\ 
{\footnotesize $\mathrm{MLP^\dagger}$ } & {\footnotesize$0.8633 \pm 0.0029$}\\ 
{\footnotesize $\mathrm{TabM^\dagger}$ } & {\footnotesize$0.8609 \pm 0.0014$}\\ 
{\footnotesize $\mathrm{TabPack}$ } & {\footnotesize$0.8575 \pm 0.0014$}\\ 
{\footnotesize $\mathrm{MLP^\dagger_{{HPE}}}$ } & {\footnotesize$0.8624 \pm 0.0013$}\\ 
{\footnotesize $\mathrm{TabPack^\dagger_{Offline}}$ } & {\footnotesize$0.8562 \pm 0.0048$}\\ 
{\footnotesize $\mathrm{TabPack_{MacBook}^\dagger}$ } & {\footnotesize$0.8588 \pm 0.0052$}\\ 
{\footnotesize $\mathrm{TabPack^\dagger}$ } & {\footnotesize$0.8623 \pm 0.0028$}\\ 
\bottomrule
\end{tabular}}

&

\topalign{
\centering
\setlength\tabcolsep{2.5pt}
\renewcommand{\arraystretch}{0.8}
\begin{tabular}{ll}

\multicolumn{2}{c}{\small{California \textdownarrow}} \\
\toprule
{\small Method} & {\small Score} \\
\midrule\\[-0.7cm]
\multicolumn{2}{c}{} \\[0.05cm]
{\footnotesize $\mathrm{MLP}$ } & {\footnotesize$0.4819 \pm 0.0023$}\\ 
{\footnotesize $\mathrm{XGBoost}$ } & {\footnotesize$0.4329 \pm 0.0011$}\\ 
{\footnotesize $\mathrm{TabICLv2}$ } & {\footnotesize$0.3976 \pm 0.0005$}\\ 
{\footnotesize $\mathrm{TabPFN\texttt{-}3}$ } & {\footnotesize$0.3787 \pm 0.0007$}\\ 
{\footnotesize $\mathrm{ModernNCA^\dagger}$ } & {\footnotesize$0.4098 \pm 0.0035$}\\ 
{\footnotesize $\mathrm{RealMLP}$ } & {\footnotesize$0.4063 \pm 0.0011$}\\ 
{\footnotesize $\mathrm{TabM}$ } & {\footnotesize$0.4324 \pm 0.0026$}\\ 
{\footnotesize $\mathrm{MLP^\dagger}$ } & {\footnotesize$0.4376 \pm 0.0022$}\\ 
{\footnotesize $\mathrm{TabM^\dagger}$ } & {\footnotesize$0.4028 \pm 0.0019$}\\ 
{\footnotesize $\mathrm{TabPack}$ } & {\footnotesize$0.4725 \pm 0.0014$}\\ 
{\footnotesize $\mathrm{MLP^\dagger_{{HPE}}}$ } & {\footnotesize$0.4233 \pm 0.0024$}\\ 
{\footnotesize $\mathrm{TabPack^\dagger_{Offline}}$ } & {\footnotesize$0.4241 \pm 0.0025$}\\ 
{\footnotesize $\mathrm{TabPack_{MacBook}^\dagger}$ } & {\footnotesize$0.4171 \pm 0.0011$}\\ 
{\footnotesize $\mathrm{TabPack^\dagger}$ } & {\footnotesize$0.4176 \pm 0.0028$}\\ 
\bottomrule
\end{tabular}}

\\
\topalign{
\centering
\setlength\tabcolsep{2.5pt}
\renewcommand{\arraystretch}{0.8}
\begin{tabular}{ll}

\multicolumn{2}{c}{\small{House \textdownarrow}} \\
\toprule
{\small Method} & {\small Score} \\
\midrule\\[-0.7cm]
\multicolumn{2}{c}{} \\[0.05cm]
{\footnotesize $\mathrm{MLP}$ } & {\footnotesize$30387 \pm 214$}\\ 
{\footnotesize $\mathrm{XGBoost}$ } & {\footnotesize$31381 \pm 60$}\\ 
{\footnotesize $\mathrm{TabICLv2}$ } & {\footnotesize$27958 \pm 87$}\\ 
{\footnotesize $\mathrm{TabPFN\texttt{-}3}$ } & {\footnotesize$26913 \pm 75$}\\ 
{\footnotesize $\mathrm{ModernNCA^\dagger}$ } & {\footnotesize$30713 \pm 399$}\\ 
{\footnotesize $\mathrm{RealMLP}$ } & {\footnotesize$30077 \pm 138$}\\ 
{\footnotesize $\mathrm{TabM}$ } & {\footnotesize$30390 \pm 124$}\\ 
{\footnotesize $\mathrm{MLP^\dagger}$ } & {\footnotesize$29851 \pm 411$}\\ 
{\footnotesize $\mathrm{TabM^\dagger}$ } & {\footnotesize$30151 \pm 218$}\\ 
{\footnotesize $\mathrm{TabPack}$ } & {\footnotesize$29721 \pm 91$}\\ 
{\footnotesize $\mathrm{MLP^\dagger_{{HPE}}}$ } & {\footnotesize$29611 \pm 198$}\\ 
{\footnotesize $\mathrm{TabPack^\dagger_{Offline}}$ } & {\footnotesize$29896 \pm 179$}\\ 
{\footnotesize $\mathrm{TabPack_{MacBook}^\dagger}$ } & {\footnotesize$29477 \pm 162$}\\ 
{\footnotesize $\mathrm{TabPack^\dagger}$ } & {\footnotesize$29549 \pm 175$}\\ 
\bottomrule
\end{tabular}}

&

\topalign{
\centering
\setlength\tabcolsep{2.5pt}
\renewcommand{\arraystretch}{0.8}
\begin{tabular}{ll}

\multicolumn{2}{c}{\small{Sberbank-Housing \textdownarrow}} \\
\toprule
{\small Method} & {\small Score} \\
\midrule\\[-0.7cm]
\multicolumn{2}{c}{} \\[0.05cm]
{\footnotesize $\mathrm{MLP}$ } & {\footnotesize$0.2608 \pm 0.0149$}\\ 
{\footnotesize $\mathrm{XGBoost}$ } & {\footnotesize$0.2407 \pm 0.0004$}\\ 
{\footnotesize $\mathrm{TabICLv2}$ } & {\footnotesize$0.3137 \pm 0.0107$}\\ 
{\footnotesize $\mathrm{TabPFN\texttt{-}3}$ } & {\footnotesize$0.2288 \pm 0.0013$}\\ 
{\footnotesize $\mathrm{ModernNCA^\dagger}$ } & {\footnotesize$0.2379 \pm 0.0047$}\\ 
{\footnotesize $\mathrm{RealMLP}$ } & {\footnotesize$0.2302 \pm 0.0012$}\\ 
{\footnotesize $\mathrm{TabM}$ } & {\footnotesize$0.2424 \pm 0.0033$}\\ 
{\footnotesize $\mathrm{MLP^\dagger}$ } & {\footnotesize$0.2481 \pm 0.0038$}\\ 
{\footnotesize $\mathrm{TabM^\dagger}$ } & {\footnotesize$0.2343 \pm 0.0013$}\\ 
{\footnotesize $\mathrm{TabPack}$ } & {\footnotesize$0.2470 \pm 0.0018$}\\ 
{\footnotesize $\mathrm{MLP^\dagger_{{HPE}}}$ } & {\footnotesize$0.2345 \pm 0.0010$}\\ 
{\footnotesize $\mathrm{TabPack^\dagger_{Offline}}$ } & {\footnotesize$0.2339 \pm 0.0023$}\\ 
{\footnotesize $\mathrm{TabPack_{MacBook}^\dagger}$ } & {\footnotesize$0.2309 \pm 0.0017$}\\ 
{\footnotesize $\mathrm{TabPack^\dagger}$ } & {\footnotesize$0.2302 \pm 0.0014$}\\ 
\bottomrule
\end{tabular}}

\\
\topalign{
\centering
\setlength\tabcolsep{2.5pt}
\renewcommand{\arraystretch}{0.8}
\begin{tabular}{ll}

\multicolumn{2}{c}{\small{Adult \textuparrow}} \\
\toprule
{\small Method} & {\small Score} \\
\midrule\\[-0.7cm]
\multicolumn{2}{c}{} \\[0.05cm]
{\footnotesize $\mathrm{MLP}$ } & {\footnotesize$0.8561 \pm 0.0012$}\\ 
{\footnotesize $\mathrm{XGBoost}$ } & {\footnotesize$0.8709 \pm 0.0007$}\\ 
{\footnotesize $\mathrm{TabICLv2}$ } & {\footnotesize$0.8699 \pm 0.0022$}\\ 
{\footnotesize $\mathrm{TabPFN\texttt{-}3}$ } & {\footnotesize$0.8636 \pm 0.0002$}\\ 
{\footnotesize $\mathrm{ModernNCA^\dagger}$ } & {\footnotesize$0.8687 \pm 0.0011$}\\ 
{\footnotesize $\mathrm{RealMLP}$ } & {\footnotesize$0.8717 \pm 0.0014$}\\ 
{\footnotesize $\mathrm{TabM}$ } & {\footnotesize$0.8577 \pm 0.0007$}\\ 
{\footnotesize $\mathrm{MLP^\dagger}$ } & {\footnotesize$0.8692 \pm 0.0018$}\\ 
{\footnotesize $\mathrm{TabM^\dagger}$ } & {\footnotesize$0.8674 \pm 0.0010$}\\ 
{\footnotesize $\mathrm{TabPack}$ } & {\footnotesize$0.8572 \pm 0.0011$}\\ 
{\footnotesize $\mathrm{MLP^\dagger_{{HPE}}}$ } & {\footnotesize$0.8671 \pm 0.0012$}\\ 
{\footnotesize $\mathrm{TabPack^\dagger_{Offline}}$ } & {\footnotesize$0.8680 \pm 0.0023$}\\ 
{\footnotesize $\mathrm{TabPack_{MacBook}^\dagger}$ } & {\footnotesize$0.8694 \pm 0.0019$}\\ 
{\footnotesize $\mathrm{TabPack^\dagger}$ } & {\footnotesize$0.8693 \pm 0.0007$}\\ 
\bottomrule
\end{tabular}}

&

\topalign{
\centering
\setlength\tabcolsep{2.5pt}
\renewcommand{\arraystretch}{0.8}
\begin{tabular}{ll}

\multicolumn{2}{c}{\small{Diamond \textdownarrow}} \\
\toprule
{\small Method} & {\small Score} \\
\midrule\\[-0.7cm]
\multicolumn{2}{c}{} \\[0.05cm]
{\footnotesize $\mathrm{MLP}$ } & {\footnotesize$0.1359 \pm 0.0012$}\\ 
{\footnotesize $\mathrm{XGBoost}$ } & {\footnotesize$0.1337 \pm 0.0004$}\\ 
{\footnotesize $\mathrm{TabICLv2}$ } & {\footnotesize$0.1249 \pm 0.0002$}\\ 
{\footnotesize $\mathrm{TabPFN\texttt{-}3}$ } & {\footnotesize$0.1246 \pm 0.0002$}\\ 
{\footnotesize $\mathrm{ModernNCA^\dagger}$ } & {\footnotesize$0.1331 \pm 0.0018$}\\ 
{\footnotesize $\mathrm{RealMLP}$ } & {\footnotesize$0.1308 \pm 0.0006$}\\ 
{\footnotesize $\mathrm{TabM}$ } & {\footnotesize$0.1315 \pm 0.0010$}\\ 
{\footnotesize $\mathrm{MLP^\dagger}$ } & {\footnotesize$0.1343 \pm 0.0008$}\\ 
{\footnotesize $\mathrm{TabM^\dagger}$ } & {\footnotesize$0.1306 \pm 0.0007$}\\ 
{\footnotesize $\mathrm{TabPack}$ } & {\footnotesize$0.1331 \pm 0.0003$}\\ 
{\footnotesize $\mathrm{MLP^\dagger_{{HPE}}}$ } & {\footnotesize$0.1307 \pm 0.0005$}\\ 
{\footnotesize $\mathrm{TabPack^\dagger_{Offline}}$ } & {\footnotesize$0.1309 \pm 0.0003$}\\ 
{\footnotesize $\mathrm{TabPack_{MacBook}^\dagger}$ } & {\footnotesize$0.1303 \pm 0.0003$}\\ 
{\footnotesize $\mathrm{TabPack^\dagger}$ } & {\footnotesize$0.1307 \pm 0.0007$}\\ 
\bottomrule
\end{tabular}}

\\
\topalign{
\centering
\setlength\tabcolsep{2.5pt}
\renewcommand{\arraystretch}{0.8}
\begin{tabular}{ll}

\multicolumn{2}{c}{\small{Otto \textuparrow}} \\
\toprule
{\small Method} & {\small Score} \\
\midrule\\[-0.7cm]
\multicolumn{2}{c}{} \\[0.05cm]
{\footnotesize $\mathrm{MLP}$ } & {\footnotesize$0.8218 \pm 0.0029$}\\ 
{\footnotesize $\mathrm{XGBoost}$ } & {\footnotesize$0.8301 \pm 0.0016$}\\ 
{\footnotesize $\mathrm{TabICLv2}$ } & {\footnotesize$0.8434 \pm 0.0012$}\\ 
{\footnotesize $\mathrm{TabPFN\texttt{-}3}$ } & {\footnotesize$0.8405 \pm 0.0008$}\\ 
{\footnotesize $\mathrm{ModernNCA^\dagger}$ } & {\footnotesize$0.8298 \pm 0.0022$}\\ 
{\footnotesize $\mathrm{RealMLP}$ } & {\footnotesize$0.8287 \pm 0.0007$}\\ 
{\footnotesize $\mathrm{TabM}$ } & {\footnotesize$0.8278 \pm 0.0012$}\\ 
{\footnotesize $\mathrm{MLP^\dagger}$ } & {\footnotesize$0.8237 \pm 0.0023$}\\ 
{\footnotesize $\mathrm{TabM^\dagger}$ } & {\footnotesize$0.8332 \pm 0.0021$}\\ 
{\footnotesize $\mathrm{TabPack}$ } & {\footnotesize$0.8238 \pm 0.0012$}\\ 
{\footnotesize $\mathrm{MLP^\dagger_{{HPE}}}$ } & {\footnotesize$0.8241 \pm 0.0018$}\\ 
{\footnotesize $\mathrm{TabPack^\dagger_{Offline}}$ } & {\footnotesize$0.8248 \pm 0.0020$}\\ 
{\footnotesize $\mathrm{TabPack_{MacBook}^\dagger}$ } & {\footnotesize$0.8292 \pm 0.0017$}\\ 
{\footnotesize $\mathrm{TabPack^\dagger}$ } & {\footnotesize$0.8281 \pm 0.0017$}\\ 
\bottomrule
\end{tabular}}

&

\topalign{
\centering
\setlength\tabcolsep{2.5pt}
\renewcommand{\arraystretch}{0.8}
\begin{tabular}{ll}

\multicolumn{2}{c}{\small{Higgs-Small \textuparrow}} \\
\toprule
{\small Method} & {\small Score} \\
\midrule\\[-0.7cm]
\multicolumn{2}{c}{} \\[0.05cm]
{\footnotesize $\mathrm{MLP}$ } & {\footnotesize$0.7267 \pm 0.0013$}\\ 
{\footnotesize $\mathrm{XGBoost}$ } & {\footnotesize$0.7271 \pm 0.0009$}\\ 
{\footnotesize $\mathrm{TabICLv2}$ } & {\footnotesize$0.7355 \pm 0.0009$}\\ 
{\footnotesize $\mathrm{TabPFN\texttt{-}3}$ } & {\footnotesize$0.7391 \pm 0.0008$}\\ 
{\footnotesize $\mathrm{ModernNCA^\dagger}$ } & {\footnotesize$0.7310 \pm 0.0014$}\\ 
{\footnotesize $\mathrm{RealMLP}$ } & {\footnotesize$0.7311 \pm 0.0014$}\\ 
{\footnotesize $\mathrm{TabM}$ } & {\footnotesize$0.7409 \pm 0.0029$}\\ 
{\footnotesize $\mathrm{MLP^\dagger}$ } & {\footnotesize$0.7279 \pm 0.0007$}\\ 
{\footnotesize $\mathrm{TabM^\dagger}$ } & {\footnotesize$0.7352 \pm 0.0008$}\\ 
{\footnotesize $\mathrm{TabPack}$ } & {\footnotesize$0.7288 \pm 0.0021$}\\ 
{\footnotesize $\mathrm{MLP^\dagger_{{HPE}}}$ } & {\footnotesize$0.7313 \pm 0.0008$}\\ 
{\footnotesize $\mathrm{TabPack^\dagger_{Offline}}$ } & {\footnotesize$0.7309 \pm 0.0007$}\\ 
{\footnotesize $\mathrm{TabPack_{MacBook}^\dagger}$ } & {\footnotesize$0.7316 \pm 0.0018$}\\ 
{\footnotesize $\mathrm{TabPack^\dagger}$ } & {\footnotesize$0.7317 \pm 0.0010$}\\ 
\bottomrule
\end{tabular}}

\\
\topalign{
\centering
\setlength\tabcolsep{2.5pt}
\renewcommand{\arraystretch}{0.8}
\begin{tabular}{ll}

\multicolumn{2}{c}{\small{Black-Friday \textdownarrow}} \\
\toprule
{\small Method} & {\small Score} \\
\midrule\\[-0.7cm]
\multicolumn{2}{c}{} \\[0.05cm]
{\footnotesize $\mathrm{MLP}$ } & {\footnotesize$0.6927 \pm 0.0006$}\\ 
{\footnotesize $\mathrm{XGBoost}$ } & {\footnotesize$0.6808 \pm 0.0001$}\\ 
{\footnotesize $\mathrm{TabICLv2}$ } & {\footnotesize$0.6919 \pm 0.0002$}\\ 
{\footnotesize $\mathrm{TabPFN\texttt{-}3}$ } & {\footnotesize$0.7026 \pm 0.0028$}\\ 
{\footnotesize $\mathrm{ModernNCA^\dagger}$ } & {\footnotesize$0.6861 \pm 0.0006$}\\ 
{\footnotesize $\mathrm{RealMLP}$ } & {\footnotesize$0.6781 \pm 0.0004$}\\ 
{\footnotesize $\mathrm{TabM}$ } & {\footnotesize$0.6846 \pm 0.0003$}\\ 
{\footnotesize $\mathrm{MLP^\dagger}$ } & {\footnotesize$0.6816 \pm 0.0006$}\\ 
{\footnotesize $\mathrm{TabM^\dagger}$ } & {\footnotesize$0.6766 \pm 0.0006$}\\ 
{\footnotesize $\mathrm{TabPack}$ } & {\footnotesize$0.6877 \pm 0.0003$}\\ 
{\footnotesize $\mathrm{MLP^\dagger_{{HPE}}}$ } & {\footnotesize$0.6802 \pm 0.0003$}\\ 
{\footnotesize $\mathrm{TabPack^\dagger_{Offline}}$ } & {\footnotesize$0.6797 \pm 0.0005$}\\ 
{\footnotesize $\mathrm{TabPack_{MacBook}^\dagger}$ } & {\footnotesize$0.6783 \pm 0.0001$}\\ 
{\footnotesize $\mathrm{TabPack^\dagger}$ } & {\footnotesize$0.6784 \pm 0.0002$}\\ 
\bottomrule
\end{tabular}}

&

\topalign{
\centering
\setlength\tabcolsep{2.5pt}
\renewcommand{\arraystretch}{0.8}
\begin{tabular}{ll}

\multicolumn{2}{c}{\small{Ecom-Offers \textuparrow}} \\
\toprule
{\small Method} & {\small Score} \\
\midrule\\[-0.7cm]
\multicolumn{2}{c}{} \\[0.05cm]
{\footnotesize $\mathrm{MLP}$ } & {\footnotesize$0.6011 \pm 0.0025$}\\ 
{\footnotesize $\mathrm{XGBoost}$ } & {\footnotesize$0.5741 \pm 0.0055$}\\ 
{\footnotesize $\mathrm{TabICLv2}$ } & {\footnotesize$0.5974 \pm 0.0020$}\\ 
{\footnotesize $\mathrm{TabPFN\texttt{-}3}$ } & {\footnotesize$0.6343 \pm 0.0009$}\\ 
{\footnotesize $\mathrm{ModernNCA^\dagger}$ } & {\footnotesize$0.5831 \pm 0.0019$}\\ 
{\footnotesize $\mathrm{RealMLP}$ } & {\footnotesize$0.5806 \pm 0.0145$}\\ 
{\footnotesize $\mathrm{TabM}$ } & {\footnotesize$0.6016 \pm 0.0004$}\\ 
{\footnotesize $\mathrm{MLP^\dagger}$ } & {\footnotesize$0.5993 \pm 0.0008$}\\ 
{\footnotesize $\mathrm{TabM^\dagger}$ } & {\footnotesize$0.5984 \pm 0.0010$}\\ 
{\footnotesize $\mathrm{TabPack}$ } & {\footnotesize$0.6018 \pm 0.0012$}\\ 
{\footnotesize $\mathrm{MLP^\dagger_{{HPE}}}$ } & {\footnotesize$0.5986 \pm 0.0017$}\\ 
{\footnotesize $\mathrm{TabPack^\dagger_{Offline}}$ } & {\footnotesize$0.5987 \pm 0.0009$}\\ 
{\footnotesize $\mathrm{TabPack_{MacBook}^\dagger}$ } & {\footnotesize$0.5990 \pm 0.0013$}\\ 
{\footnotesize $\mathrm{TabPack^\dagger}$ } & {\footnotesize$0.5989 \pm 0.0017$}\\ 
\bottomrule
\end{tabular}}

\\
\topalign{
\centering
\setlength\tabcolsep{2.5pt}
\renewcommand{\arraystretch}{0.8}
\begin{tabular}{ll}

\multicolumn{2}{c}{\small{Maps-Routing \textdownarrow}} \\
\toprule
{\small Method} & {\small Score} \\
\midrule\\[-0.7cm]
\multicolumn{2}{c}{} \\[0.05cm]
{\footnotesize $\mathrm{MLP}$ } & {\footnotesize$0.1622 \pm 0.0001$}\\ 
{\footnotesize $\mathrm{XGBoost}$ } & {\footnotesize$0.1618 \pm 0.0000$}\\ 
{\footnotesize $\mathrm{TabICLv2}$ } & {\footnotesize$0.1657 \pm 0.0002$}\\ 
{\footnotesize $\mathrm{TabPFN\texttt{-}3}$ } & {\footnotesize$0.1643 \pm 0.0002$}\\ 
{\footnotesize $\mathrm{ModernNCA^\dagger}$ } & {\footnotesize$0.1628 \pm 0.0000$}\\ 
{\footnotesize $\mathrm{RealMLP}$ } & {\footnotesize$0.1609 \pm 0.0001$}\\ 
{\footnotesize $\mathrm{TabM}$ } & {\footnotesize$0.1611 \pm 0.0001$}\\ 
{\footnotesize $\mathrm{MLP^\dagger}$ } & {\footnotesize$0.1612 \pm 0.0001$}\\ 
{\footnotesize $\mathrm{TabM^\dagger}$ } & {\footnotesize$0.1606 \pm 0.0002$}\\ 
{\footnotesize $\mathrm{TabPack}$ } & {\footnotesize$0.1612 \pm 0.0000$}\\ 
{\footnotesize $\mathrm{MLP^\dagger_{{HPE}}}$ } & {\footnotesize$0.1607 \pm 0.0001$}\\ 
{\footnotesize $\mathrm{TabPack^\dagger_{Offline}}$ } & {\footnotesize$0.1610 \pm 0.0001$}\\ 
{\footnotesize $\mathrm{TabPack_{MacBook}^\dagger}$ } & {\footnotesize$0.1604 \pm 0.0001$}\\ 
{\footnotesize $\mathrm{TabPack^\dagger}$ } & {\footnotesize$0.1605 \pm 0.0001$}\\ 
\bottomrule
\end{tabular}}

&

\topalign{
\centering
\setlength\tabcolsep{2.5pt}
\renewcommand{\arraystretch}{0.8}
\begin{tabular}{ll}

\multicolumn{2}{c}{\small{Homesite-Insurance \textuparrow}} \\
\toprule
{\small Method} & {\small Score} \\
\midrule\\[-0.7cm]
\multicolumn{2}{c}{} \\[0.05cm]
{\footnotesize $\mathrm{MLP}$ } & {\footnotesize$0.9515 \pm 0.0007$}\\ 
{\footnotesize $\mathrm{XGBoost}$ } & {\footnotesize$0.9606 \pm 0.0001$}\\ 
{\footnotesize $\mathrm{TabICLv2}$ } & {\footnotesize$0.9470 \pm 0.0046$}\\ 
{\footnotesize $\mathrm{TabPFN\texttt{-}3}$ } & {\footnotesize$0.9568 \pm 0.0015$}\\ 
{\footnotesize $\mathrm{ModernNCA^\dagger}$ } & {\footnotesize$0.9629 \pm 0.0003$}\\ 
{\footnotesize $\mathrm{RealMLP}$ } & {\footnotesize$0.9600 \pm 0.0015$}\\ 
{\footnotesize $\mathrm{TabM}$ } & {\footnotesize$0.9652 \pm 0.0004$}\\ 
{\footnotesize $\mathrm{MLP^\dagger}$ } & {\footnotesize$0.9627 \pm 0.0007$}\\ 
{\footnotesize $\mathrm{TabM^\dagger}$ } & {\footnotesize$0.9644 \pm 0.0007$}\\ 
{\footnotesize $\mathrm{TabPack}$ } & {\footnotesize$0.9518 \pm 0.0002$}\\ 
{\footnotesize $\mathrm{MLP^\dagger_{{HPE}}}$ } & {\footnotesize$0.9638 \pm 0.0004$}\\ 
{\footnotesize $\mathrm{TabPack^\dagger_{Offline}}$ } & {\footnotesize$0.9640 \pm 0.0004$}\\ 
{\footnotesize $\mathrm{TabPack_{MacBook}^\dagger}$ } & {\footnotesize$0.9642 \pm 0.0001$}\\ 
{\footnotesize $\mathrm{TabPack^\dagger}$ } & {\footnotesize$0.9644 \pm 0.0003$}\\ 
\bottomrule
\end{tabular}}

\\
\topalign{
\centering
\setlength\tabcolsep{2.5pt}
\renewcommand{\arraystretch}{0.8}
\begin{tabular}{ll}

\multicolumn{2}{c}{\small{Cooking-Time \textdownarrow}} \\
\toprule
{\small Method} & {\small Score} \\
\midrule\\[-0.7cm]
\multicolumn{2}{c}{} \\[0.05cm]
{\footnotesize $\mathrm{MLP}$ } & {\footnotesize$0.4824 \pm 0.0002$}\\ 
{\footnotesize $\mathrm{XGBoost}$ } & {\footnotesize$0.4824 \pm 0.0001$}\\ 
{\footnotesize $\mathrm{TabICLv2}$ } & {\footnotesize$0.4851 \pm 0.0004$}\\ 
{\footnotesize $\mathrm{TabPFN\texttt{-}3}$ } & {\footnotesize$0.4844 \pm 0.0004$}\\ 
{\footnotesize $\mathrm{ModernNCA^\dagger}$ } & {\footnotesize$0.4813 \pm 0.0003$}\\ 
{\footnotesize $\mathrm{RealMLP}$ } & {\footnotesize$0.4809 \pm 0.0005$}\\ 
{\footnotesize $\mathrm{TabM}$ } & {\footnotesize$0.4807 \pm 0.0002$}\\ 
{\footnotesize $\mathrm{MLP^\dagger}$ } & {\footnotesize$0.4810 \pm 0.0001$}\\ 
{\footnotesize $\mathrm{TabM^\dagger}$ } & {\footnotesize$0.4802 \pm 0.0002$}\\ 
{\footnotesize $\mathrm{TabPack}$ } & {\footnotesize$0.4810 \pm 0.0000$}\\ 
{\footnotesize $\mathrm{MLP^\dagger_{{HPE}}}$ } & {\footnotesize$0.4795 \pm 0.0003$}\\ 
{\footnotesize $\mathrm{TabPack^\dagger_{Offline}}$ } & {\footnotesize$0.4798 \pm 0.0002$}\\ 
{\footnotesize $\mathrm{TabPack_{MacBook}^\dagger}$ } & {\footnotesize$0.4793 \pm 0.0002$}\\ 
{\footnotesize $\mathrm{TabPack^\dagger}$ } & {\footnotesize$0.4792 \pm 0.0002$}\\ 
\bottomrule
\end{tabular}}

&

\topalign{
\centering
\setlength\tabcolsep{2.5pt}
\renewcommand{\arraystretch}{0.8}
\begin{tabular}{ll}

\multicolumn{2}{c}{\small{Homecredit-Default \textuparrow}} \\
\toprule
{\small Method} & {\small Score} \\
\midrule\\[-0.7cm]
\multicolumn{2}{c}{} \\[0.05cm]
{\footnotesize $\mathrm{MLP}$ } & {\footnotesize$0.8551 \pm 0.0009$}\\ 
{\footnotesize $\mathrm{XGBoost}$ } & {\footnotesize$0.8675 \pm 0.0003$}\\ 
{\footnotesize $\mathrm{TabICLv2}$ } & {\footnotesize$0.8454 \pm 0.0019$}\\ 
{\footnotesize $\mathrm{TabPFN\texttt{-}3}$ } & {\footnotesize$0.8501 \pm 0.0014$}\\ 
{\footnotesize $\mathrm{ModernNCA^\dagger}$ } & {\footnotesize$0.8526 \pm 0.0019$}\\ 
{\footnotesize $\mathrm{RealMLP}$ } & {\footnotesize$0.8509 \pm 0.0023$}\\ 
{\footnotesize $\mathrm{TabM}$ } & {\footnotesize$0.8613 \pm 0.0011$}\\ 
{\footnotesize $\mathrm{MLP^\dagger}$ } & {\footnotesize$0.8601 \pm 0.0004$}\\ 
{\footnotesize $\mathrm{TabM^\dagger}$ } & {\footnotesize$0.8642 \pm 0.0006$}\\ 
{\footnotesize $\mathrm{TabPack}$ } & {\footnotesize$0.8535 \pm 0.0003$}\\ 
{\footnotesize $\mathrm{MLP^\dagger_{{HPE}}}$ } & {\footnotesize$0.8615 \pm 0.0015$}\\ 
{\footnotesize $\mathrm{TabPack^\dagger_{Offline}}$ } & {\footnotesize$0.8617 \pm 0.0010$}\\ 
{\footnotesize $\mathrm{TabPack_{MacBook}^\dagger}$ } & {\footnotesize$0.8611 \pm 0.0007$}\\ 
{\footnotesize $\mathrm{TabPack^\dagger}$ } & {\footnotesize$0.8615 \pm 0.0006$}\\ 
\bottomrule
\end{tabular}}

\\
\topalign{
\centering
\setlength\tabcolsep{2.5pt}
\renewcommand{\arraystretch}{0.8}
\begin{tabular}{ll}

\multicolumn{2}{c}{\small{Delivery-Eta \textdownarrow}} \\
\toprule
{\small Method} & {\small Score} \\
\midrule\\[-0.7cm]
\multicolumn{2}{c}{} \\[0.05cm]
{\footnotesize $\mathrm{MLP}$ } & {\footnotesize$0.5476 \pm 0.0005$}\\ 
{\footnotesize $\mathrm{XGBoost}$ } & {\footnotesize$0.5458 \pm 0.0001$}\\ 
{\footnotesize $\mathrm{TabICLv2}$ } & {\footnotesize$0.5537 \pm 0.0003$}\\ 
{\footnotesize $\mathrm{TabPFN\texttt{-}3}$ } & {\footnotesize$0.5545 \pm 0.0009$}\\ 
{\footnotesize $\mathrm{ModernNCA^\dagger}$ } & {\footnotesize$0.5508 \pm 0.0003$}\\ 
{\footnotesize $\mathrm{RealMLP}$ } & {\footnotesize$0.5468 \pm 0.0009$}\\ 
{\footnotesize $\mathrm{TabM}$ } & {\footnotesize$0.5483 \pm 0.0002$}\\ 
{\footnotesize $\mathrm{MLP^\dagger}$ } & {\footnotesize$0.5499 \pm 0.0008$}\\ 
{\footnotesize $\mathrm{TabM^\dagger}$ } & {\footnotesize$0.5464 \pm 0.0004$}\\ 
{\footnotesize $\mathrm{TabPack}$ } & {\footnotesize$0.5502 \pm 0.0004$}\\ 
{\footnotesize $\mathrm{MLP^\dagger_{{HPE}}}$ } & {\footnotesize$0.5490 \pm 0.0009$}\\ 
{\footnotesize $\mathrm{TabPack^\dagger_{Offline}}$ } & {\footnotesize$0.5486 \pm 0.0009$}\\ 
{\footnotesize $\mathrm{TabPack_{MacBook}^\dagger}$ } & {\footnotesize$0.5474 \pm 0.0007$}\\ 
{\footnotesize $\mathrm{TabPack^\dagger}$ } & {\footnotesize$0.5477 \pm 0.0005$}\\ 
\bottomrule
\end{tabular}}

&

\topalign{
\centering
\setlength\tabcolsep{2.5pt}
\renewcommand{\arraystretch}{0.8}
\begin{tabular}{ll}

\multicolumn{2}{c}{\small{Weather \textdownarrow}} \\
\toprule
{\small Method} & {\small Score} \\
\midrule\\[-0.7cm]
\multicolumn{2}{c}{} \\[0.05cm]
{\footnotesize $\mathrm{MLP}$ } & {\footnotesize$1.5010 \pm 0.0037$}\\ 
{\footnotesize $\mathrm{XGBoost}$ } & {\footnotesize$1.4694 \pm 0.0005$}\\ 
{\footnotesize $\mathrm{TabICLv2}$ } & {\footnotesize$1.4690 \pm 0.0029$}\\ 
{\footnotesize $\mathrm{TabPFN\texttt{-}3}$ } & {\footnotesize$1.4597 \pm 0.0011$}\\ 
{\footnotesize $\mathrm{ModernNCA^\dagger}$ } & {\footnotesize$1.5007 \pm 0.0028$}\\ 
{\footnotesize $\mathrm{RealMLP}$ } & {\footnotesize$1.4398 \pm 0.0016$}\\ 
{\footnotesize $\mathrm{TabM}$ } & {\footnotesize$1.4651 \pm 0.0042$}\\ 
{\footnotesize $\mathrm{MLP^\dagger}$ } & {\footnotesize$1.4915 \pm 0.0020$}\\ 
{\footnotesize $\mathrm{TabM^\dagger}$ } & {\footnotesize$1.4491 \pm 0.0034$}\\ 
{\footnotesize $\mathrm{TabPack}$ } & {\footnotesize$1.4742 \pm 0.0010$}\\ 
{\footnotesize $\mathrm{MLP^\dagger_{{HPE}}}$ } & {\footnotesize$1.4607 \pm 0.0021$}\\ 
{\footnotesize $\mathrm{TabPack^\dagger_{Offline}}$ } & {\footnotesize$1.4659 \pm 0.0040$}\\ 
{\footnotesize $\mathrm{TabPack_{MacBook}^\dagger}$ } & {\footnotesize$1.4436 \pm 0.0013$}\\ 
{\footnotesize $\mathrm{TabPack^\dagger}$ } & {\footnotesize$1.4430 \pm 0.0007$}\\ 
\bottomrule
\end{tabular}}

\\
\topalign{
\centering
\setlength\tabcolsep{2.5pt}
\renewcommand{\arraystretch}{0.8}
\begin{tabular}{ll}

\multicolumn{2}{c}{\small{Microsoft \textdownarrow}} \\
\toprule
{\small Method} & {\small Score} \\
\midrule\\[-0.7cm]
\multicolumn{2}{c}{} \\[0.05cm]
{\footnotesize $\mathrm{MLP}$ } & {\footnotesize$0.7455 \pm 0.0003$}\\ 
{\footnotesize $\mathrm{XGBoost}$ } & {\footnotesize$0.7412 \pm 0.0001$}\\ 
{\footnotesize $\mathrm{TabICLv2}$ } & {\footnotesize$0.7457 \pm 0.0002$}\\ 
{\footnotesize $\mathrm{TabPFN\texttt{-}3}$ } & {\footnotesize$0.7785 \pm 0.0026$}\\ 
{\footnotesize $\mathrm{ModernNCA^\dagger}$ } & {\footnotesize$0.7443 \pm 0.0005$}\\ 
{\footnotesize $\mathrm{RealMLP}$ } & {\footnotesize$0.7446 \pm 0.0009$}\\ 
{\footnotesize $\mathrm{TabM}$ } & {\footnotesize$0.7422 \pm 0.0004$}\\ 
{\footnotesize $\mathrm{MLP^\dagger}$ } & {\footnotesize$0.7447 \pm 0.0002$}\\ 
{\footnotesize $\mathrm{TabM^\dagger}$ } & {\footnotesize$0.7407 \pm 0.0002$}\\ 
{\footnotesize $\mathrm{TabPack}$ } & {\footnotesize$0.7423 \pm 0.0002$}\\ 
{\footnotesize $\mathrm{MLP^\dagger_{{HPE}}}$ } & {\footnotesize$0.7421 \pm 0.0005$}\\ 
{\footnotesize $\mathrm{TabPack^\dagger_{Offline}}$ } & {\footnotesize$0.7424 \pm 0.0002$}\\ 
{\footnotesize $\mathrm{TabPack_{MacBook}^\dagger}$ } & {\footnotesize$0.7395 \pm 0.0002$}\\ 
{\footnotesize $\mathrm{TabPack^\dagger}$ } & {\footnotesize$0.7395 \pm 0.0002$}\\ 
\bottomrule
\end{tabular}}

& 

\\

\end{longtable}
\label{A:tab:per-dataset-results}

\end{document}